\pdfoutput=1

\documentclass[11pt]{article}

\usepackage[final]{acl}

\usepackage{times}
\usepackage{latexsym}

\usepackage[T1]{fontenc}

\usepackage[utf8]{inputenc}

\usepackage{microtype}

\usepackage{inconsolata}

\usepackage{graphicx}
\usepackage{booktabs}

\usepackage{placeins}

\usepackage{longtable}
\usepackage{float}  

\usepackage[table,xcdraw]{xcolor}
\usepackage{soul}

\newcommand{\hlSocial}[1]{\begingroup\sethlcolor{blue!15}\textbf{\hl{#1}}\endgroup\textsuperscript{SD}}
\newcommand{\hlFear}[1]{\begingroup\sethlcolor{purple!15}\textbf{\hl{#1}}\endgroup\textsuperscript{F}}
\newcommand{\hlMoral}[1]{\begingroup\sethlcolor{green!18}\textbf{\hl{#1}}\endgroup\textsuperscript{MF}}

\usepackage{tabularx}
\usepackage{array}
\usepackage{multicol}
\usepackage{multirow}

\usepackage{tcolorbox}
\usepackage{colortbl}
\tcbuselibrary{most}
\usepackage{rotating}

\newcommand{\yo}[1]{\textcolor{pink}{#1}}
\usepackage{placeins}
\usepackage{float}

\newtcolorbox{prompt}{
  enhanced,
  colback=gray!10,
  colframe=gray!50,
  arc=0mm,
  boxrule=0.5pt,
  left=5mm,
  right=5mm,
  top=2mm,
  bottom=2mm,
  fonttitle=\bfseries,
  coltitle=black,
  breakable,
  fontupper=\footnotesize\ttfamily,
      overlay={%
        \ifcase\tcbsegmentstate
        \or%
        \else%
        \fi%
    }
}
\usepackage{cleveref}
\usepackage{comment}

\title{Automatic Evaluation of Mental Health Stigma in Online Communication}

\author{
\fontsize{10.5}{12.5}\selectfont
\textbf{Naomi Baes\textsuperscript{1,2}\thanks{These authors contributed equally and share first authorship.}},
\textbf{Jemima Kang\textsuperscript{3}\footnotemark[1]},
\textbf{Nick Haslam\textsuperscript{1}},
\textbf{Chris Groot\textsuperscript{1}},
\textbf{Alsa Wu\textsuperscript{1}}
\\[-0.1ex]
\fontsize{10.5}{12.5}\selectfont
\textbf{Luc Raszewski\textsuperscript{3}},
\textbf{Yulia Otmakhova\textsuperscript{3,1}}
\\[0.6ex]
\normalfont\fontsize{9}{11}\selectfont
\textsuperscript{1}Melbourne School of Psychological Sciences, The University of Melbourne
\\
\normalfont\fontsize{9}{11}\selectfont
\textsuperscript{2}Department of General and Computational Linguistics, University of Tübingen
\\
\normalfont\fontsize{9}{11}\selectfont
\textsuperscript{3}School of Computing and Information Systems, The University of Melbourne
\\[0.6ex]
\normalfont\fontsize{8.5}{10.5}\selectfont
\texttt{\{nbaes,jemima.kang.1,nhaslam,grootc,alsa.wu,l.raszewski,y.otmakhova\}@unimelb.edu.au}
}

\begin{document}
\maketitle
\begin{abstract}

\yo{Warning: Due to the nature of this work, some content in this paper may be sensitive, offensive, or disturbing.}

Mental health stigma has profoundly harmful impacts but its complexity makes it difficult to evaluate. Stigma may involve explicit derogation, but also subtler forms of blame, fear, paternalistic pity, social distancing, structural exclusion, and discrimination. We introduce a theory-grounded benchmark for automatic evaluation of mental health stigma in online communication, consisting of naturally occurring online news and social media text annotated with a fine-grained taxonomy of stigma across multiple mental health conditions. Our annotation framework comprises a binary stigma-detection task and a multi-level taxonomy covering (i) stigma mode, (ii) domain, and (iii) specific components of certain forms of stigma. We apply this framework to texts mentioning six mental health conditions and evaluate large language models alongside stigma-related classifiers for detecting sentiment, toxicity, and hate speech. Results show that mental health stigma is not well captured by models trained to detect these neighboring constructs, and that LLMs often overpredict stigma unless given explicit operational rules - mirroring the importance of decision rules in human annotation.\footnote{We release the publicly available part of benchmark, annotations, prototypical exemplars of stigma and code at: \url{https://github.com/jemimakang/mh_stigma}.} 
\end{abstract}

\section{Introduction}


The language of mental health stigma is profoundly harmful. It can reproduce stereotypes, justify social exclusion, discourage help-seeking, and shape how people with mental illness are perceived and treated in everyday life \cite{sartorius2007stigma, thornicroft2022lancet}. Online communication is an important site for studying these processes: news and social media circulate public representations of mental illness on an unprecedented scale, making them central channels through which stigma may be expressed, amplified, contested, or normalized \cite{groot2020}. However, despite growing interest in harmful online communication, mental health stigma remains difficult to evaluate automatically.
\begin{figure}[t]
\centering
\fbox{
\begin{minipage}{0.95\columnwidth}
\small
\textbf{Source:} Paraphrased benchmark example (Reddit) \\
\textbf{Domain/mode:} Public stigma; expressed/enacted \\
\rule{\linewidth}{0.4pt}
\textit{``I don't want XXX with her history of alcohol problems 
\hlSocial{anywhere near my family}. 
I'm \hlFear{worried about the distress she could cause} 
to my children, and I think she would 
\hlMoral{make it her goal to hurt us}.''}
\end{minipage}
}
\caption{Paraphrased benchmark example illustrating multiple stigma components: \small
\textsuperscript{SD}~Social distancing; 
\textsuperscript{F}~Fear; 
\textsuperscript{MF}~Moral or character flaw.}
\label{fig:exemplar-stigma}
\end{figure}

A central challenge is that stigma is not reducible to related NLP constructs such as negative sentiment, toxicity, or hate speech. Stigmatizing passages may contain explicit insults or hostility, but they may also involve subtler forms of devaluation, such as portraying people with mental illness as dangerous, unpredictable, blameworthy, incompetent, pitiable, or socially undesirable. Other cases involve perceived or experienced stigma, structural disadvantage, or the internalization of public stereotypes. These forms may not be globally negative in sentiment or toxic in conventional terms, meaning that models trained to detect similar constructs may miss important expressions of stigma while over-identifying general negativity as stigma. Figure~\ref{fig:exemplar-stigma} illustrates this challenge: a single passage can express multiple theoretically distinct stigma components, including social distancing, fear, and moralized attribution, without necessarily fitting neatly into a toxicity or sentiment category.

Stigma is also difficult to study at scale. Traditional approaches in psychology often rely on self-report measures or 
elicited judgments, which are valuable for measuring attitudes but less suited to tracking how stigma is transmitted in naturally occurring communication \cite{corrigan2003attribution,griffiths2008predictors,taylor1981scaling}. Existing text-based approaches offer a partial solution but have often focused on specific mental illnesses, narrow stigma components, or neighboring constructs such as hate speech and sentiment. This creates both a practical and conceptual need for benchmarks that evaluate mental health stigma as it appears in public discourse: across conditions and across explicit and subtle forms of stigmatization. Because mental health stigma is multidimensional, useful stigma detection requires more than a binary label; systems should ideally distinguish whether a passage involves fear, blame, social distancing, perceived discrimination, or other theoretically meaningful components \cite{link2001conceptualizing,thornicroft2022lancet}.

In the present study, we introduce a theory-informed evaluation benchmark that complements existing stigma-detection resources by examining naturally occurring news and social media text across multiple mental health conditions. The benchmark comprises a binary stigma-detection task and a multi-level taxonomy for fine-grained stigma annotation. The taxonomy distinguishes stigma mode, stigma domain, public-stigma subfactors (types), and public-stigma components (subtypes). It uses \citet{link2001conceptualizing}'s account of stigma as a conceptual anchor for deciding whether a passage is in scope, distinguishes expressed/enacted from perceived/experienced stigma, delineates public, structural, associative, and self-stigma, and gives greatest granularity to public stigma by separating cognitive stereotypes, affective reactions, and behavioral discrimination. We apply it to online news and Reddit posts mentioning six mental health conditions: \emph{ADHD}, \emph{alcoholism}, \emph{autism spectrum disorder}, \emph{borderline personality disorder}, \emph{depression}, and \emph{schizophrenia}.


We use the resulting benchmark to evaluate whether large language models (LLMs) can detect mental health stigma, whether explicit operational rules improve detection, whether stigma is captured by neighboring constructs such as sentiment, toxicity, and hate speech, and whether performance varies across conditions. Our experiments compare multiple prompt settings for LLMs, including definitions, examples, decision rules, and stigma-type information, alongside classifiers for related constructs. Results show that LLMs often achieve high recall but only because they substantially overpredict stigma, especially when decision rules are underspecified. Adding explicit operational rules reduces this bias, suggesting that theoretically precise operationalization is essential for reliable automatic stigma evaluation. 
Our contributions are: 
\begin{enumerate}
    \item A theory-grounded annotation framework for fine-grained mental health stigma annotation.
    \item A benchmark of naturally occurring online news and social media text annotated for stigma across six mental health conditions.
    \item An evaluation showing that mental health stigma is not well captured by LLMs or sentiment, toxicity, hate-speech classifiers.
\end{enumerate}

\section{Related work} 


\subsection{Stigma as a multidimensional construct} 
Stigma is a multidimensional social process that in the context of mental health, can be expressed cognitively (stereotypes, causal attributions), affectively (fear, disgust, pity, or ambivalence), or behaviorally (social distancing, coercion, or discriminatory treatment) \cite{corrigan2003attribution,thornicroft2022lancet}. Frameworks further distinguish that stigma may be expressed by others (public), embedded in institutions (structural), extended to people beyond those who experience mental ill health (associative), or internalized by people with lived experience (self-stigma)  \cite{thornicroft2022lancet}.

This multidimensionality challenges automatic detection, as stigma is not reducible to negative sentiment, toxicity, or overt abuse \cite{corrigan2002understanding}. Each dimension of stigma leaves distinct textual signatures. For example, blame through moralized claims of weakness or irresponsibility \cite{corrigan2003attribution,griffiths2008predictors}, dangerousness through associations with violence or criminality \cite{corrigan2003attribution,taylor1981scaling}, and social distancing through statements of avoidance, exclusion, or unwillingness to interact with people with mental illness \cite{corrigan2003attribution,link2001conceptualizing}. Immutability or "prognostic pessimism" may appear through claims that a condition is permanent, hopeless, or untreatable \cite{ritsher2003internalized,day2007measuring}. Stigma can also be less overtly hostile, including paternalistic pity or condescension, or expressed through first-person accounts of being feared, excluded, mocked, or discriminated against \cite{corrigan2003attribution,fox2018conceptualizing}. 

Stigma is also not expressed uniformly across mental health conditions \cite{price2022}. Goulden et al. (\citeyear{goulden2011}) found that UK newspaper coverage improved for some conditions, including \emph{depression}, \emph{anxiety}, \emph{bipolar disorder}, and \emph{eating disorders}, while general references to other mental illnesses (\emph{schizophrenia}, \emph{personality disorders}) were more covered in the context of ``bad news''. More recent work likewise finds that \emph{alcohol addiction} and \emph{schizophrenia} are among the most stigmatized mental disorders in public perception \cite{schomerus2011stigma, dambrun2024social}. These findings support the need to evaluate stigma across multiple mental health conditions rather than treating mental illness as a single homogeneous target.

\FloatBarrier
\subsection{Detection of mental health stigma} 
A growing body of NLP work has examined stigma in online health communication through lexicon, topic-modeling, sentiment, and framing approaches in social media and news \cite{Mittal2023,pavlova2022mental,wanniarachchi2023hate,CHEN2022100061}. Supervised classifiers have also been developed for specific stigma forms, including stigma regarding Alzheimer's disease \citep{oscar2017machine}, schizophrenia \citep{jilka2022identifying}, stigmatizing uses of psychiatric terminology \citep{moriceau2022automatic}, stigma toward people who use substances in Reddit posts \citep{giorgi2024lived}, and stigma in healthcare discussions \citep{straton2020stigma}. Recent work has also used few-shot LLM prompting to detect substance use stigma on Reddit \citep{bouzoubaa2024words,bouzoubaa2025phenotypes}. While these studies demonstrate that stigma is computationally detectable, they are tied to specific conditions and text types. 

Most of the existing studies also focus on detection of stigma presence, without drilling into its types. Closest to our work, \citet{meng2025stigma} annotate human-chatbot interviews about a vignette character with depression, labeled with seven attribution-model categories and a non-stigmatizing class. 
However, the corpus is based on chatbot interactions and elicited interview responses rather than naturally occurring online communication, and does not treat stigma as multi-dimensional construct, as we do.

We address these gaps by constructing a multi-level, fine-grained stigma evaluation framework which targets multiple mental health conditions and spans two domains -- news and social media.

\section{Method}

\subsection{Stigma taxonomy}
\label{sec:stigma_taxonomy}


We developed a hierarchical stigma taxonomy to support fine-grained annotation of mental health stigma in naturally occurring text, which conceptually draws on \citet{link2001conceptualizing}'s account of stigma. 
The taxonomy includes stigma \textit{mode}, \textit{domain}, and \textit{public-domain subtypes}. Mode distinguishes stigma that is expressed or enacted on behalf of ``the stigmatizer'' 
from stigma that is perceived or experienced on the part of the ``stigmatized'' (i.e., people with lived or living experience of mental illness) \cite{fox2018conceptualizing}. Domain distinguishes public, structural, associative, and self-stigma, following the Lancet Commission taxonomy \cite{thornicroft2022lancet}. 
As the present study focuses on public online communication, the taxonomy gives greatest granularity to public stigma, which we divide into cognitive stereotypes, affective reactions, and behavioral discrimination \cite{corrigan2003attribution}. Cognitive labels capture beliefs such as dangerousness, blame, incompetence, or immutability; affective labels capture emotions such as fear, anger, disgust, pity, or ambivalence; and behavioral labels capture discriminatory orientations such as social distancing or mocking. 

The annotation framework operationalizes this taxonomy in two stages: binary stigma detection and multi-level fine-grained annotation (see \Cref{sec:annotation}). 
Table~\ref{tab:label_support} summarizes the taxonomy structure; see Appendix~\ref{appendix:appendix_stigmataxonomy} for definitions and examples.

\subsection{Online communication} 

We use two corpora: online news from the News on the Web (NOW) corpus and Reddit posts. From NOW, we extracted U.S. news articles from 2010--2025 containing variants of the six target mental health terms. From Reddit, we extracted posts from 2012--2022 containing one target term and at least 30 words, excluding posts mentioning multiple target conditions. 
Preprocessing procedures and descriptive counts are reported in Appendix~\ref{sec:appendix_preprocessing}.





\subsection{Mental health conditions}

Our target set comprised six mental health conditions selected to balance theoretical coverage, corpus frequency, and stigma variation: \emph{ADHD}, \emph{alcoholism}, \emph{autism}, \emph{borderline personality disorder}, \emph{depression}, and \emph{schizophrenia}. Candidates were drawn from major HiTOP spectra \cite{kotov2017hierarchical} and refined in consultation with domain experts. For each condition, we constructed surface variants and manually inspected searches to exclude spurious matches, such as non-psychiatric uses of ``BPD''. See Appendix~\ref{sec:appendix_preprocessing} for search terms.



\subsection{Annotation and benchmark construction} 
\label{sec:annotation}

The benchmark was constructed through a four-round iterative annotation process. Annotation was guided by the taxonomy described in Section~\ref{sec:stigma_taxonomy} and by a codebook containing category definitions, decision rules, and examples\footnote{Available at \url{https://github.com/jemimakang/mh_stigma}} (see \Cref{app:annotation_procedure} for more details). The annotation team comprised two psychology professors with expertise in mental health stigma and two psychology PhD students focusing on stigma (the latter compensated at USD 40/hour across 50 hours).  

The annotators were shown each text alongside its target mental health condition and first judged whether the text expressed mental health stigma. When stigma was present, they assigned the relevant stigma mode, domain, and fine-grained public-stigma labels, and marked spans of text containing stigma. 
Because initial random sampling yielded relatively few stigma-present texts, later rounds used targeted sampling, LLM-assisted filtering, and model-disagreement sampling to increase the number and diversity of stigma-present and challenging cases. As a result, label proportions in the benchmark are not estimates of stigma prevalence.


\noindent\textbf{Annotation agreement.} Disagreement was concentrated around the stigma-presence threshold and, among stigma-present items, around stigma mode. Pooled across rounds, agreement on stigma presence was moderate (exact agreement = .66; AC1 = .40), as was agreement on stigma mode (exact = .61; AC1 = .54). On items that all annotators judged stigma-present, agreement was high for domain (macro pairwise AC1 = .82) and public-stigma subfactors (macro pairwise AC1 = .89), although this subset was small and some labels were rare. Full round-level and pairwise results are in~\Cref{app:human_agreement}.

All samples with disagreement were adjudicated through a discussion. Disagreement analyses after earlier rounds were also used to refine the operational rules, especially for distinguishing texts that display stigma from texts that merely discuss stigma, and for distinguishing expressed/enacted from perceived/experienced stigma. Further details of the annotation instructions, rounds, adjudication process, and guideline revisions are reported in Appendix~\ref{app:annotation_procedure}.

\subsubsection{Benchmark composition} 

Table~\ref{tab:label_support} summarizes the composition of the final benchmark. The benchmark contains 470 texts, of which 87 were annotated as stigma-present. 
Expressed/enacted was the most common stigma mode, and public stigma the most common domain. Within public stigma, cognitive components were assigned far more often than affective or behavioral ones. The cognitive component pity/condescension (patronizing appraisal of a person as helpless, childlike, inferior, or incapable) was assigned to eight texts, while the affective component pity was assigned to none. Affective pity (explicitly expressed emotional response of pity, sorrow, or sympathy) is retained in the taxonomy on theoretical grounds but excluded from category-level evaluation. Additionally, bullying/mocking was a particularly present 'other' category during the annotation process therefore, it was added to the category-level evaluation.

\begin{table}[t]
\centering
\footnotesize
\setlength{\tabcolsep}{4pt}
\renewcommand{\arraystretch}{0.95}
\begin{tabular*}{\columnwidth}{@{}l@{\extracolsep{\fill}}r@{}}
\toprule
\textbf{Label} & \textbf{N} \\
\midrule
\textbf{Binary stigma}\\
Stigma absent & 383 \\
Stigma present & 87 \\
\hspace*{1em}\textbf{Mode} & \\
\hspace*{2em}Expressed/enacted & 49 \\
\hspace*{2em}Perceived/experienced & 15 \\
\hspace*{2em}Both & 23 \\
\hspace*{1em}\textbf{Domain} & \\
\hspace*{2em}\textit{Public} & 62 \\
\hspace*{3em}Cognitive & 54 \\
\hspace*{4em}Moral or character flaw & 14 \\
\hspace*{4em}Incompetence & 14 \\
\hspace*{4em}Uncontrollable & 11 \\
\hspace*{4em}Trivialization/minimization & 9 \\
\hspace*{4em}Pity/condescension & 8 \\
\hspace*{4em}Dangerousness & 7 \\
\hspace*{4em}Immutability & 5 \\
\hspace*{4em}Segregation endorsement & 4 \\
\hspace*{4em}Controllable & 3 \\
\hspace*{4em}Personal responsibility/blame & 2 \\
\hspace*{4em}Contamination & 1 \\
\hspace*{4em}Coercion & 0 \\
\hspace*{3em}Affective & 9 \\
\hspace*{4em}Anger/irritability/frustration & 4 \\
\hspace*{4em}Disgust & 4 \\
\hspace*{4em}Fear & 2 \\
\hspace*{4em}Ambivalence & 0 \\
\hspace*{4em}Pity & 0 \\
\hspace*{3em}Behavioral & 5 \\
\hspace*{4em}Social distancing & 1 \\
\hspace*{4em}Bullying/mocking & 4 \\
\hspace*{2em}\textit{Self} & 24 \\
\hspace*{2em}\textit{Structural} & 6 \\
\hspace*{2em}\textit{Associative} & 1 \\
\bottomrule
\end{tabular*}
\caption{Label support in the final benchmark (470 texts). Source and condition breakdowns are given in Table~\ref{tab:benchmark_composition_expanded}.}
\label{tab:label_support}
\end{table}

\subsection{Stigma prototype exemplars}
\label{sec:prototype_exemplars}

To support our experiments, we constructed a separate prototype exemplar set for the stigma categories. These examples were not part of the annotated benchmark; they were standalone examples designed to represent each stigma type clearly. For each category, we created stigma-present and stigma-absent examples. Where existing stigma scale items directly captured a category, we adapted their wording minimally; where no scale items were available, examples were composed from the formal category definitions. Stigma-absent examples were generally constructed by negating or reframing the corresponding stigma-present item, and domain experts expanded categories with fewer than ten suitable examples. The prototype dataset was used to provide demonstration examples in LLM prompts and to support embedding similarity experiments. See details in~\Cref{appendix:prototype_exemplars}.




\section{Automatic detection of stigma presence and dimensions} 

In this section, we apply our benchmark to test the ability of large language models (LLMs) to detect mental health stigma and classify its types across the dimensions defined in \Cref{sec:stigma_taxonomy}. 


\subsection{Experiment details}

\noindent\textbf{Tasks and data.} The number of samples evaluated depends on the task: for \textit{binary classification of stigma presence}, we use all annotated samples in the benchmark (N=470). We use only stigmatizing samples (n=87) for two \textit{multi-label classification tasks}: predicting \textbf{stigma mode} (3 classes - \textit{Expressed/Enacted}, \textit{Perceived/Experienced}, \textit{Both}) and \textbf{stigma domain} (4 classes - \textit{Public}, \textit{Structural}, \textit{Associative}, \textit{Self}). Finally, since we annotate stigma subfactors (\textit{Cognitive}, \textit{Affective}, \textit{Behavioral}), as well as their components, only in texts with public domain stigma, we use only 62 samples for corresponding multi-label classifications.\footnote{We adjust the number of classes for fine-grained components to reflect their occurrence in annotated data (11 classes for Cognitive type, 3 for Affective type, and 2 for Behavioral type, see Panel D of \Cref{tab:benchmark_composition_expanded}).}

\noindent\textbf{Prompts.} We use the following experiment conditions to examine if LLMs are able to predict stigma and its types when provided their definitions, examples, additional rules (if available), or just given the corresponding classes:

\begin{itemize}
    \item \textbf{baseline}: we only specify the task and the classes being predicted.
    \item \textbf{definition}: in addition to the instruction, the prompt contains definitions of classes. For stigma presence, we use the definition proposed by \citet{link2001conceptualizing} (``Stigma exists when elements of labeling, stereotyping, separation, status loss, and discrimination occur together.''); theory-driven definitions for stigma modes, domains, subfactors, and components are in \Cref{tab:appendix_stigma_taxonomy}.
    \item \textbf{examples}: in addition to the instruction, the prompt contains examples drawn from our prototypes set (1 for each class).
    \item \textbf{definition+examples}: the prompt contains both the definition and examples of stigma.
    \item \textbf{{definition+examples+rule}}: in addition to the above, the prompt contains the operational rules that were used by annotators to clarify harder cases of stigma presence or mode (see \Cref{sec:annotation}).
\end{itemize}

All prompts were developed in consultation with domain experts and iteratively refined.

\noindent\textbf{Models.} We evaluate representative models from four major providers: Anthropic's Claude Sonnet 4.6, OpenAI's GPT 5.5, Google's Gemini 3 Flash, and DeepSeek V4 Flash (model identifiers and inference settings are in \Cref{app:model_details}). We choose models covering a range of sizes and capabilities: a distilled model (Gemini 3 Flash), a mid-tier model (Sonnet 4.6), and frontier reasoning models (GPT 5.5 and DeepSeek V4 Flash).

For the binary stigma prediction, we additionally LoRA-tune a Qwen2.5-7B model\footnote{Model from: https://huggingface.co/Qwen/Qwen2.5-7B; scripts for LoRA provided on \url{https://github.com/jemimakang/mh_stigma}} with a classifier head on the mental health stigma dataset by \citet{meng2025stigma}. To adapt the dataset of our binary stigma presence detection task, we convert all labels representing stigmatizing classes into a single ``stigma\_present'' label. We do not use \citet{meng2025stigma}'s dataset for classifying fine-grained stigma types as their taxonomy is different from ours.

\begin{table*}[th!]
\centering
\small
\setlength{\tabcolsep}{5pt}
\begin{tabular}{llcccccccc}
\toprule
& & & \multicolumn{3}{c}{\textbf{Binary}} & \multicolumn{3}{c}{\textbf{Macro}} & \\
\cmidrule(lr){4-6} \cmidrule(lr){7-9}
\textbf{Strategy} & \textbf{Model} & \textbf{Acc} & \textbf{P} & \textbf{R} & \textbf{F1} & \textbf{P} & \textbf{R} & \textbf{F1} & \textbf{Bias} \\
\midrule
 
\multirow{4}{*}{\texttt{baseline}}
 & Claude Sonnet 4.6 & 0.666 & 0.310 & 0.655 & 0.421 & 0.602 & 0.662 & 0.593 & \cellcolor{red!54}0.206 \\
 & Deepseek v4 Flash & 0.704 & 0.268 & 0.345 & 0.302 & 0.554 & 0.565 & 0.557 & \cellcolor{red!14}0.053 \\
 & Gemini 3 Flash    & 0.734 & 0.386 & 0.736 & \textbf{0.506} & \textbf{\textit{0.655}} & 0.735 & \textbf{0.662} & \cellcolor{red!44}0.168 \\
 & GPT-5.5           & 0.566 & 0.281 & 0.862 & 0.424 & 0.611 & 0.680 & 0.538 & \cellcolor{red!100}0.383 \\
\midrule
 
\multirow{4}{*}{\texttt{definition}}
 & Claude Sonnet 4.6 & 0.698 & 0.315 & 0.540 & 0.398 & 0.595 & 0.637 & 0.598 & \cellcolor{red!34}0.132 \\
 & Deepseek v4 Flash & 0.626 & 0.269 & 0.598 & 0.371 & 0.572 & 0.615 & 0.552 & \cellcolor{red!59}0.226 \\
 & Gemini 3 Flash    & 0.734 & 0.378 & 0.678 & 0.486 & 0.645 & 0.712 & 0.653 & \cellcolor{red!38}0.147 \\
 & GPT-5.5           & 0.691 & 0.342 & 0.724 & 0.465 & 0.629 & 0.704 & 0.624 & \cellcolor{red!54}0.206 \\
\midrule
 
\multirow{4}{*}{\texttt{examples}}
 & Claude Sonnet 4.6 & 0.723 & 0.348 & 0.563 & 0.430 & 0.616 & 0.662 & 0.624 & \cellcolor{red!30}0.115 \\
 & Deepseek v4 Flash & 0.651 & 0.282 & 0.575 & 0.379 & 0.578 & 0.622 & 0.568 & \cellcolor{red!50}0.191 \\
 & Gemini 3 Flash    & 0.668 & 0.327 & 0.747 & 0.455 & 0.623 & 0.699 & 0.608 & \cellcolor{red!62}0.238 \\
 & GPT-5.5           & 0.649 & 0.332 & \textbf{\textit{0.885}} & 0.483 & 0.645 & \textbf{\textit{0.740}} & 0.609 & \cellcolor{red!81}0.309 \\
\midrule
 
\multirow{4}{*}{\texttt{examples+definition}}
 & Claude Sonnet 4.6 & 0.732 & 0.353 & 0.540 & 0.427 & 0.617 & 0.658 & 0.626 & \cellcolor{red!26}0.098 \\
 & Deepseek v4 Flash & 0.640 & 0.272 & 0.563 & 0.367 & 0.571 & 0.611 & 0.558 & \cellcolor{red!52}0.198 \\
 & Gemini 3 Flash    & 0.677 & 0.328 & 0.713 & 0.449 & 0.620 & 0.691 & 0.610 & \cellcolor{red!57}0.217 \\
 & GPT-5.5           & 0.664 & 0.330 & 0.793 & 0.466 & 0.631 & 0.714 & 0.610 & \cellcolor{red!68}0.260 \\
\midrule
 
\multirow{4}{*}{\texttt{examples+definition+rule}}
 & Claude Sonnet 4.6 & \textbf{\textit{0.781}} & \textbf{\textit{0.397}} & 0.356 & 0.376 & 0.627 & 0.617 & 0.621 & \cellcolor{blue!5}$-$0.019 \\
 & Deepseek v4 Flash & 0.706 & 0.314 & 0.494 & 0.384 & 0.591 & 0.624 & 0.596 & \cellcolor{red!28}0.106 \\
 & Gemini 3 Flash    & 0.732 & 0.356 & 0.552 & 0.432 & 0.620 & 0.662 & 0.628 & \cellcolor{red!27}0.102 \\
 & GPT-5.5           & 0.666 & 0.325 & 0.747 & 0.453 & 0.622 & 0.697 & 0.606 & \cellcolor{red!63}0.240 \\

 
\bottomrule
\end{tabular}
\caption{Stigma presence detection results (accuracy, as well binary and macro-averaged precision, recall, and F1) across experimental settings and models. Binary metrics are reported in terms of the positive class (``stigma present''). Bias = model's predicted positive rate $-$ gold stigma-positive rate;  positive values (red) = overprediction, negative values (blue) = underprediction. Bracketed values are 95\% confidence intervals obtained from bootstrapping (2000 resamples). Best results across all models and conditions are in \textbf{bold}; \textit{\textbf{bold italics}}  means that the difference with the next-best result for a particular metric is also statistically significant (per-instance McNemar test, $\chi^2$ with continuity correction, p < 0.05). }
\label{tab:stigma_presence_results}
\end{table*}

\subsection{Results}

\subsubsection{Stigma presence classification}
\label{sec:stigma_presence}

The results for binary classification of stigma presence are shown in \Cref{tab:stigma_presence_results} (full results together with confidence intervals are in \Cref{tab:stigma_presence_cis}). In addition to the binary metrics -- to evaluate the correctness of positive class (``stigma present'') -- and macro-averaged metrics we report the difference in the model's stigma prediction rate and the actual presence rate (``Bias''). The latter metric highlights whether the model tends to flag stigma more aggressively or more conservatively than expected.

Surprisingly, Gemini with the baseline prompt achieves the highest binary F1, showing the best balance between precision and recall for detecting stigma. Its high recall comes with substantial over-prediction, however, and the advantage does not extend to finer-grained prediction of stigma types. This suggests that the model errs on the side of caution, flagging stigma when uncertain, though our experiments do not establish whether it results from model safeguards or from another mechanism. Definitions do not appear to help, while some models (GPT 5.5, and to a lesser extent Sonnet 4.6) predict stigma more accurately when given examples, but this shows an association between the examples and the predictions rather than acquisition of the underlying conceptual distinctions. Most importantly, the operational rules substantially increase the precision of stigma detection and overall accuracy, especially for Claude Sonnet 4.6, and reduce the rate of over-prediction.

\subsubsection{Transfer from existing stigma data}

\begin{table*}[th]
\centering
\small
\setlength{\tabcolsep}{4pt}
\begin{tabular}{llcccccccc}
\toprule
& & & \multicolumn{3}{c}{\textbf{Binary}} & \multicolumn{3}{c}{\textbf{Macro}} & \\
\cmidrule(lr){4-6} \cmidrule(lr){7-9}
\textbf{Test set} & \textbf{Model} & \textbf{Acc} & \textbf{P} & \textbf{R} & \textbf{F1} & \textbf{P} & \textbf{R} & \textbf{F1} & \textbf{Bias} \\
\midrule
Meng's & Meng's & 0.917 & 0.908 & 0.915 & 0.911 & 0.916 & 0.917 & 0.917 & \cellcolor{red!1}0.004 \\
\midrule
\multirow{3}{*}{Ours}
 & Meng's (zero-shot) & 0.306 & 0.165 & 0.678 & 0.266 & 0.459 & 0.450 & 0.304 & \cellcolor{red!100}0.574 \\
 & Meng's + ours$^\dagger$ & 0.755 & 0.321 & 0.287 & 0.303 & 0.581 & 0.574 & 0.577 & \cellcolor{blue!4}$-$0.019 \\
 & Ours$^\dagger$ & 0.753 & 0.321 & 0.299 & 0.310 & 0.582 & 0.578 & 0.580 & \cellcolor{blue!3}$-$0.013 \\
\bottomrule
\end{tabular}
\caption{LoRA-tuned classifier results for stigma presence (Qwen2.5-7B). $^\dagger$5-fold cross-validation on our benchmark; metrics are computed over the pooled out-of-fold predictions from all five folds. Bias = predicted positive rate $-$ gold positive rate; positive values (red) = overprediction, negative values (blue) = underprediction.}
\label{tab:finetuned}
\end{table*}

The model fine-tuned on the adapted \citet{meng2025stigma} dataset transferred poorly to our benchmark (\Cref{tab:finetuned}). While it performs strongly on the original test set, on our benchmark its accuracy, precision and recall drop dramatically. The model fine-tuned on \citet{meng2025stigma} severely overpredicts stigma (bias $+$0.574), and despite a large number of false alarms (binary precision 0.165), it still fails to detect all stigmatizing texts. Since the model was trained on chatbot interactions, in which participants answer questions that directly probe their attitudes towards a person with a mental illness, it may have learned superficial cues, such as evaluative talk about mental illness, that do not generalize to the subtler and more diverse ways stigma appears in naturally occurring news and Reddit posts, where most texts mention a condition are not stigmatizing.

Continued training on our benchmark (5-fold cross-validation) removes the over-prediction tendency, but performance is no better than that of a model trained on our data alone. Training on the data of \citet{meng2025stigma} thus provides no measurable benefit for our task.


\subsubsection{Predicting stigma modes, domains, types and subtypes}

We next evaluate whether models can classify stigma beyond binary presence. Table~\ref{tab:merged} reports performance for stigma mode, domain, public-stigma type, and cognitive components (full details together with confidence intervals in \Cref{app:detailed_scores}). These tasks differ in granularity: mode and domain capture higher-level distinctions, whereas type and component prediction require models to distinguish finer-grained public-stigma labels. Moreover, the number of samples for each label varies drastically, and thus the results for finer-grained categories (for example, performance for affective or behavioral components) should be interpreted with caution.

\begin{table*}[ht]
\centering
\small
\begin{tabular}{llcccccccc}
\toprule
& & \multicolumn{2}{c}{\textbf{Mode}} & \multicolumn{2}{c}{\textbf{Domain}} & \multicolumn{2}{c}{\textbf{Type}} & \multicolumn{2}{c}{\textbf{Cognitive}} \\
\cmidrule(lr){3-4} \cmidrule(lr){5-6} \cmidrule(lr){7-8} \cmidrule(lr){9-10}
\textbf{Strategy} & \textbf{Model} & \textbf{Acc} & \textbf{F1} & \textbf{Acc} & \textbf{F1} & \textbf{Acc} & \textbf{F1} & \textbf{Acc} & \textbf{F1} \\
\midrule
 
\multirow{4}{*}{\texttt{baseline}}
 & Claude Sonnet 4.6 & 0.563 & 0.809 & 0.529 & 0.706 & 0.386 & 0.528 & 0.14 & 0.49 \\
 & Deepseek v4 Flash & 0.575 & 0.759 & 0.483 & 0.532 & 0.351 & 0.441 & 0.12 & 0.36 \\
 & Gemini 3 Flash    & 0.575 & 0.792 & 0.517 & 0.559 & 0.632 & 0.426 & 0.18 & 0.48 \\
 & GPT-5.5           & 0.621 & 0.837 & 0.713 & 0.854 & 0.596 & \textbf{0.562} & 0.24 & 0.50 \\
\midrule
 
\multirow{4}{*}{\texttt{definition}}
 & Claude Sonnet 4.6 & 0.540 & 0.798 & 0.563 & 0.662 & 0.474 & 0.473 & 0.22 & 0.48 \\
 & Deepseek v4 Flash & 0.632 & 0.826 & 0.575 & 0.572 & 0.456 & 0.489 & 0.18 & 0.38 \\
 & Gemini 3 Flash    & 0.586 & 0.787 & 0.598 & 0.601 & 0.667 & 0.494 & 0.16 & 0.42 \\
 & GPT-5.5           & 0.609 & 0.836 & 0.724 & 0.780 & 0.561 & 0.528 & 0.22 & \textbf{0.56} \\
\midrule
 
\multirow{4}{*}{\texttt{examples}}
 & Claude Sonnet 4.6 & 0.552 & 0.804 & 0.552 & 0.696 & 0.509 & 0.535 & 0.16 & 0.44 \\
 & Deepseek v4 Flash & 0.575 & 0.773 & 0.632 & 0.656 & 0.491 & 0.479 & 0.10 & 0.29 \\
 & Gemini 3 Flash    & 0.575 & 0.806 & 0.621 & 0.595 & \textbf{0.737} & 0.530 & 0.20 & 0.49 \\
 & GPT-5.5           & 0.701 & 0.878 & \textbf{0.759} & \textbf{0.877} & 0.596 & 0.501 & 0.20 & 0.44 \\
\midrule
 
 
\multirow{4}{*}{\texttt{examples+definition}}
 & Claude Sonnet 4.6 & 0.517 & 0.761 & 0.575 & 0.598 & 0.526 & 0.523 & \textbf{0.24} & 0.46 \\
 & Deepseek v4 Flash & 0.598 & 0.801 & 0.632 & 0.642 & 0.456 & 0.503 & 0.14 & 0.37 \\
 & Gemini 3 Flash    & 0.552 & 0.741 & 0.598 & 0.581 & 0.719 & 0.504 & 0.18 & 0.49 \\
 & GPT-5.5           & 0.690 & 0.871 & 0.747 & 0.856 & 0.614 & 0.522 & \textbf{0.24} & 0.55 \\
\midrule
 
\multirow{4}{*}{\texttt{+rules}}
 & Claude Sonnet 4.6 & 0.678 & 0.801 & \multicolumn{6}{c}{---} \\
 & Deepseek v4 Flash & 0.575 & 0.804 & \multicolumn{6}{c}{---} \\
 & Gemini 3 Flash    & 0.805 & 0.896 & \multicolumn{6}{c}{---} \\
 & GPT-5.5           & \textbf{0.816} & \textbf{0.921} & \multicolumn{6}{c}{---} \\
 
\bottomrule
\end{tabular}
\caption{Predicting stigma mode (n=87), domain (n=87), type (n=57), and subtype (for Cognitive type only, n=49) (accuracy / macro-F1). The Cognitive subtype task is multi-label, and accuracy is evaluated strictly (all labels match).}
\label{tab:merged}
\end{table*}

Across fine-grained stigma prediction tasks, different kinds of prompt information helped at different annotation levels. Operational rules substantially improved stigma-mode prediction, consistent with the human annotation process in which mode distinctions required explicit decision rules. For stigma domain and public-stigma type, examples were more useful than definitions, suggesting that broader categories may be easier to learn from concrete cases than from abstract descriptions. By contrast, definitions were more helpful for fine-grained cognitive components, where closely related stereotypes require explicit conceptual boundaries. 
Confusion matrices in Appendix~\ref{app:confusion_matrices} show that several cognitive errors involved theoretically related categories: \emph{personal responsibility} was confused with \emph{trivialization}, \emph{controllability} with \emph{personal responsibility}, \emph{segregation endorsement} with \emph{incompetence}, and \emph{dangerousness} with \emph{uncontrollability}. Affective components were also confused, but these patterns should be interpreted cautiously because the number of affective examples was small.


\section{Discussion}


\subsection{Condition-specific variation in stigma detection}



\begin{table}[t]
\centering
\footnotesize
\setlength{\tabcolsep}{3pt}
\begin{tabular}{lrrrrrrr}
\toprule
\textbf{Condition} & \textbf{N} & \textbf{Stig+} & \textbf{Acc} & \textbf{P} & \textbf{R} & \textbf{F1} & \textbf{Bias} \\
\midrule
ADHD          & 77 & 13 & .758 & .344 & .462 & .389 & \cellcolor{red!12}  .060 \\
Alcoholism    & 78 & 12 & .759 & .321 & .400 & .336 & \cellcolor{red!11}  .056 \\
Autism        & 78 & 16 & .746 & .419 & .613 & .494 & \cellcolor{red!19}  .095 \\
BPD           & 80 & 20 & .585 & .295 & .470 & .358 & \cellcolor{red!30}  .150 \\
Depression    & 79 &  4 & .911 & .400 & .600 & .416 & \cellcolor{red!10}  .048 \\
Schizophrenia & 78 & 22 & .562 & .360 & .627 & .447 & \cellcolor{red!46}  .228 \\
\bottomrule
\end{tabular}
\caption{Condition-level stigma-presence results for Claude Sonnet 4.6, averaged across the five prompting strategies. N = number of items; Stig+ = gold stigma-positive items. Bias = model's predicted positive rate $-$ gold stigma-positive rate; positive values (red) = overprediction.}
\label{tab:by_condition}
\end{table}

We also conducted exploratory condition-level analyses to examine whether model performance varied between mental health conditions, which would be obscured by evaluation on a single generic mental illness category (\Cref{tab:by_condition}, full results in \Cref{tab:by_condition_full}). Claude Sonnet 4.6 overpredicted stigma most strongly for \emph{schizophrenia}, which had the largest positive bias and lowest accuracy, suggesting that the model may overgeneralize from condition-specific cues such as risk, dangerousness, or intense clinical framing. By contrast, \emph{alcoholism} and \emph{ADHD} showed higher accuracy and lower bias (we ignore the results for depression due to very few stigma-present cases). Interestingly, introduction of operational rules reduced the bias towards overprediction of stigma for schizophrenia most strongly (\Cref{tab:by_condition_full}). These patterns suggest that both benchmark composition (where such conditions as schizophrenia are over-represented even with randomization) and model errors vary by condition, supporting evaluation across multiple mental health conditions rather than a single generic category. However, because stigma prevalence was unequal across conditions and was shaped by the sampling procedure, these estimates may be sensitive to individual examples and should not be taken as evidence of reliable differences between conditions.


\subsection{Challenges in detecting mental health stigma} 



To examine whether model errors reflected genuinely difficult annotation cases, we used an item-level difficulty variable derived from pre-adjudication annotator disagreement and confidence scores.\footnote{The difficulty level is a weighted score from the annotator disagreement and confidence scores mapped to ``difficult", ``moderate", and ``easy" categories. All gold labels were adjudicated, so ``difficult'' does not indicate uncertainty in the final benchmark label.} Model errors were higher for items that were higher in difficulty: across the 20 model--prompt predictions per item, `hard' texts had a mean/median error rate of .51/.53, compared with .36/.28 for `moderate' items and .24/.10 for `easy' items (Table~\ref{tab:error_by_difficulty}). This pattern suggests that models made errors in texts that required more human deliberation during benchmark construction, rather than in the easier cases.

\begin{table}[h]
\centering
\small
\begin{tabular}{lrr}
\toprule
\textbf{Difficulty} & \textbf{N items} & \textbf{Error rate (mean / median)} \\
\midrule
Easy & 232 & .241 / .100 \\
Moderate & 190 & .356 / .275 \\
Hard & 48 & .514 / .525 \\
\bottomrule
\end{tabular}
\caption{Model error rates by item-level annotation difficulty. Error rate is the proportion of incorrect model--prompt predictions for each text, calculated across 20 model--prompt combinations.}

\label{tab:error_by_difficulty}
\end{table}

Prompt information changed model prediction patterns in different ways. For Claude Sonnet 4.6, adding operational rules and stigma-type information made the model more conservative in predicting stigma: false positives decreased from 102 under \texttt{definition} to 47 under \texttt{definition+examples+rule}. 
However, false negatives increased from 40 to 56, indicating that it also caused the model to miss some stigma-present cases. 
These results suggest that additional prompt information does not uniformly improve stigma detection; instead, it can make some models more conservative and other more sensitive to weak or indirect stigma cues. 

Qualitative inspection suggested complementary false-positive and false-negative patterns. False positives often arose when models treated negative clinical, risk-related, or metaphorical mental health contexts as stigma despite the absence of a stigmatizing attribution, treatment, or expression toward people with mental illness; for example, a question about whether cannabis use could lead to schizophrenia was over-read as stigma. False negatives tended to involve less overt cues embedded in first-person, institutional, or non-toxic discourse, such as self-directed ``shame and guilt'' about alcoholism, concern that autistic adults may be ``left behind'' by support systems, or ADHD being dismissed as ``an excuse.'' These patterns suggest that models were sensitive to negativity present in the text but less consistent at applying the adjudicated decision rule. Full false-positive and false-negative examples are reported in Appendix~\ref{app:error_examples_full_model}. 



\subsection{Dissecting stigma} 

The manual error analysis of LLMs and the fine-tuned model suggests that models often treated negativity, toxicity, or shallow lexical cues as evidence of stigma, while struggling with cases that required applying the benchmark's operational decision rule or generalizing across situations and contexts. We therefore conducted a diagnostic analysis asking 1) whether gold stigma labels were statistically associated with classifier outputs for distinct but related constructs: general and targeted sentiment, hate speech, and toxicity; 2) whether stigma can be captured through semantic similarity; 3) whether models can learn to better distinguish stigma when given more examples. 

\textbf{Association between stigma and adjacent constructs.} The Chi-squared test of independence results did not support a reliable association between stigma labels and predicted general sentiment, $\chi^2(2)=2.56$, $p=.277$; targeted sentiment, $\chi^2(2)=2.37$, $p=.306$; or hate speech, $\chi^2(1)=0.06$, $p=.813$. Toxicity scores also showed substantial overlap between stigma-present and stigma-absent texts, confirmed by statistically insignificant point-biserial correlation ($r=0.064$, $p=.169$) (see \Cref{app:adjacent_constructs} for more details and graphs). This supports the central premise of the benchmark: mental health stigma is not reducible to neighboring constructs such as negative sentiment, toxicity, or hate speech. Stigma may involve negative sentiment or derogation, but its core features and expressions are not adequately captured 
by generic classifiers for harmful language.

\textbf{Capturing stigma through semantic similarity.} To examine if presence of stigma and its types can be captured and distinguished purely based on shallow semantics, we measure the similarity of samples in our dataset to prototypical exemplars we collected (\Cref{sec:prototype_exemplars}). We find that the semantic similarity is not enough to correctly select the closest prototype to a particular text (more details and results in \Cref{app:semantic_similarity}). Contrasting this with the next experiment, this does not reflect the quality of the prototypes, as they are useful when learning to predict these categories from examples.

\textbf{Learning finer-grained distinctions from examples.} When constructing prototypical exemplars, we aimed to represent various social situations representing stigma. Correspondingly, the results should improve when a model is exposed to more of such examples (i.e., learns to extrapolate stigma to new situations). Comparing results for 1 and 10 examples 
shows that among the examined models, only GPT 5.5 is able to do so consistently, and other models in many cases get confused.




\section{Conclusion}

We introduced a theory-grounded benchmark for mental health stigma detection in naturally occurring online news and social media text, combining binary stigma detection with a fine-grained taxonomy across multiple conditions. Three findings stand out. First, LLMs under minimal prompting achieved high recall by substantially overpredicting stigma. Second, explicit operational rules improved performance, particularly by reducing false positives. Third, model errors concentrated in the same threshold cases that required greater annotator deliberation, suggesting that reliable stigma detection depends on precise construct operationalization rather than model capability alone. These results underscore that stigma is not reducible to toxicity, hate speech, or negative sentiment, and highlight the need for stigma-specific evaluation protocols. Future work should expand the dataset to include additional conditions, different annotator perspectives such as those of people with lived experience of mental health stigma, and more examples of rarer stigma categories, enabling more reliable evaluation at the finer levels of the taxonomy.

\section*{Limitations}



This study has several limitations. First, we construct and evaluate a benchmark for stigma detection rather than measure how prevalent stigma is or how it has changed over time. Because later annotation rounds enriched the sample with stigma-present, boundary, and model-disagreement cases, the label distribution reflects benchmark construction rather than prevalence in the NOW or Reddit corpora, and should not be used to estimate population-level prevalence.

Second, our annotation framework is theory-driven, but it does not directly incorporate the perspectives of people with lived or living experience of mental illness. This is important because stigma may be perceived differently by researchers, members of the general public, clinicians, and people who have experienced mental health stigma directly. 
Prominent reports, such as the Lancet Commission on ending stigma and discrimination in mental health \cite{thornicroft2022lancet}, reviewing 216 systematic reviews have emphasized the importance of include the lived experience of people with mental health conditions. As a result, the benchmark may not capture all forms of stigma that are most salient or harmful to affected communities and it is imperative for future work to extend our approach to include the perspective of people with lived experience of mental health conditions. 

Third, our benchmark treats each target condition as a broad category. This enables comparison across six mental health conditions, but it does not capture within-condition heterogeneity. Stigma may vary depending on the gender, race, class, age, or other social position of the person being described, as well as on whether a condition is represented through diagnostic, behavioral, criminal-justice, medical, or personal-recovery frames.

Fourth, each text was presented to annotators alongside its target mental health condition. Knowledge that a particular text was related to schizophrenia, alcoholism, autism, or another condition may have shaped annotators' judgments, especially for categories such as dangerousness, blame, or social distancing. Future robustness checks could repeat a subset of annotations with target terms masked or replaced by neutral placeholders, allowing researchers to estimate how much stigma detection depends on lexical labels versus surrounding context. Although source metadata was withheld, annotators may sometimes have been able to infer whether a text came from news or Reddit from stylistic or contextual cues, such as first-person language.

Fifth, our evaluation is limited to selected large language models, prompt settings, and adjacent-construct classifiers. 
Our results should therefore not be interpreted as an exhaustive comparison of all possible stigma detection systems. Instead, they evaluate whether widely available LLMs and related harmful-language classifiers can serve as adequate proxies for mental health stigma detection, and establish a benchmark against which future dedicated systems can be tested. 



Finally, our corpora are limited to online news and Reddit. These domains are important sites of public discourse, but they do not represent all communicative settings in which mental health stigma occurs. Stigma may be expressed differently in clinical records, policy documents, television transcripts, interviews, support forums, or non-English-language media, limiting the generalizability of the benchmark beyond the domains and languages studied here.

Despite these limitations, our theory-grounded benchmark establishes a foundation for future evaluation protocols that enables the kind of large-scale monitoring necessary to track stigma prevalence, evaluate the impact of public health campaigns, and ultimately support efforts to reduce it. 

\section*{Ethical Considerations}

This work addresses a sensitive domain: mental health stigma in naturally occurring online communication. We use two sources of online text: U.S. online news from the News on the Web (NOW) corpus and Reddit posts collected via Pushshift for research and benchmark evaluation. 

While the Reddit posts are publicly available and data collection was consistent with Reddit's terms and conditions \cite{reddit2024privacy}, individual consent could not be sought from Reddit users. Therefore, we adopt relevant standards outlined by \citet{benton-etal-2017-ethical} - and following \citet{Chen_Robinson_Conway_2025} - to ensure protection for individuals who post to the Reddit communities in use.
Because Reddit posts may contain sensitive self-disclosures about mental health, additional protections include the avoidance of usernames and direct links in the paper, paraphrasing illustrative examples where appropriate, and storage of data in secure environments.
Reddit data use in the study was approved by the Human Ethics Committee (Office of Research Ethics and Integrity) at the University of Melbourne (No.2026-36618-80970-1.)

The benchmark is released in two parts. All annotations, the accompanying metadata, and the excerpt text of the news items are openly available; the excerpt text of the Reddit items is available only on application to the authors. Every item is identified by a pseudonym rather than by its original Reddit or news identifier and no usernames are distributed.

The benchmark is intended as a research and evaluation resource, not as a standalone moderation, clinical, or individual-level assessment tool. Automatic stigma detection may produce false positives, incorrectly flagging clinical discussion, self-disclosure, advocacy, or accounts of experienced stigma as harmful, as well as false negatives that miss subtle forms of stigma. There is also a risk that condition-specific detection could reinforce associations between particular diagnoses and harmfulness or social risk. Future work should incorporate lived-experience perspectives in benchmark design and evaluation.

\textbf{AI use.} We used Claude for language edits, to assist code writing, and for typesetting tables. None of the work (including citations) was generated by LLMs, and all code and tables were checked by the authors of the paper.


\section*{Acknowledgments}

This project was funded by the University of Melbourne Hallmark Research Initiative on Fighting Harmful Online Communication (seed grant 026796). We thank Label Studio Platform for providing access to the Enterprise Cloud Platform through their Academic Program. Our gratitude also extends to Associate Professor Mike Conway and Dr Jacinthe Flore for early discussions.



\bibliography{custom}

\begin{thebibliography}{32}
\providecommand{\natexlab}[1]{#1}

\bibitem[{Baes et~al.(2025)Baes, Merx, Haslam, Vylomova, and Dubossarsky}]{baes-etal-2025-lsc}
Naomi Baes, Raphael Merx, Nick Haslam, Ekaterina Vylomova, and Haim Dubossarsky. 2025.
\newblock \href {https://doi.org/10.18653/v1/2025.findings-acl.570} {{LSC}-eval: A general framework to evaluate methods for assessing dimensions of lexical semantic change using {LLM}-generated synthetic data}.
\newblock In \emph{Findings of the Association for Computational Linguistics: ACL 2025}, pages 10905--10939, Vienna, Austria. Association for Computational Linguistics.

\bibitem[{Benton et~al.(2017)Benton, Coppersmith, and Dredze}]{benton-etal-2017-ethical}
Adrian Benton, Glen Coppersmith, and Mark Dredze. 2017.
\newblock \href {https://doi.org/10.18653/v1/W17-1612} {Ethical research protocols for social media health research}.
\newblock In \emph{Proceedings of the First {ACL} Workshop on Ethics in Natural Language Processing}, pages 94--102, Valencia, Spain. Association for Computational Linguistics.

\bibitem[{Bouzoubaa et~al.(2024)Bouzoubaa, Aghakhani, and Rezapour}]{bouzoubaa2024words}
Layla Bouzoubaa, Elham Aghakhani, and Rezvaneh Rezapour. 2024.
\newblock \href {https://doi.org/10.18653/v1/2024.emnlp-main.516} {Words matter: Reducing stigma in online conversations about substance use with large language models}.
\newblock In \emph{Proceedings of the 2024 Conference on Empirical Methods in Natural Language Processing}, pages 9139--9156, Miami, FL, USA. Association for Computational Linguistics.

\bibitem[{Bouzoubaa et~al.(2025)Bouzoubaa, Aghakhani, and Rezapour}]{bouzoubaa2025phenotypes}
Layla Bouzoubaa, Elham Aghakhani, and Rezvaneh Rezapour. 2025.
\newblock \href {https://doi.org/10.1016/j.socscimed.2025.118889} {Phenotypes of stigma expressed by people who use drugs on reddit}.
\newblock \emph{Social Science \& Medicine}, 390:118889.

\bibitem[{Chen et~al.(2022)Chen, Johnny, and Conway}]{CHEN2022100061}
Annie~T. Chen, Shana Johnny, and Mike Conway. 2022.
\newblock \href {https://doi.org/10.1016/j.dadr.2022.100061} {Examining stigma relating to substance use and contextual factors in social media discussions}.
\newblock \emph{Drug and Alcohol Dependence Reports}, 3:100061.

\bibitem[{Chen et~al.(2025)Chen, Robinson, and Conway}]{Chen_Robinson_Conway_2025}
Liuliu Chen, Jo~Robinson, and Mike Conway. 2025.
\newblock \href {https://doi.org/10.1609/icwsm.v19i1.35822} {{What Do You Meme? -- Identifying Characteristics and User Perceptions of Suicide Memes in Social Media}}.
\newblock In \emph{Proceedings of the International AAAI Conference on Web and Social Media}, volume~19, pages 385--402, Copenhagen, Denmark. AAAI Press.

\bibitem[{Corrigan et~al.(2003)Corrigan, Markowitz, Watson, Rowan, and Kubiak}]{corrigan2003attribution}
Patrick Corrigan, Fred~E. Markowitz, Amy Watson, David Rowan, and Mary~Ann Kubiak. 2003.
\newblock \href {https://doi.org/10.2307/1519806} {An attribution model of public discrimination towards persons with mental illness}.
\newblock \emph{Journal of Health and Social Behavior}, 44(2):162--179.

\bibitem[{Corrigan and Watson(2002)}]{corrigan2002understanding}
Patrick Corrigan and Amy~C. Watson. 2002.
\newblock Understanding the impact of stigma on people with mental illness.
\newblock \emph{World Psychiatry}, 1(1):16--20.

\bibitem[{Dambrun et~al.(2024)Dambrun, Marinho, Mondillon, Clinchamps, Dutheil, Chakroun, Teissedre, Motak, Corman, Cuchet, Fargheot, Izaute, Daugherty, de~la Sablonni{\`e}re, Lindsay, and Szeto}]{dambrun2024social}
Michael Dambrun, Ga{\'e}tan Marinho, Laurie Mondillon, Ma{\'e}lys Clinchamps, Fr{\'e}d{\'e}ric Dutheil, Nadia Chakroun, Fr{\'e}d{\'e}rique Teissedre, Ladislav Motak, Maya Corman, Isabelle Cuchet, Juliette Fargheot, Marie Izaute, Julia Daugherty, Roxane de~la Sablonni{\`e}re, Brittany~L. Lindsay, and Andrew C.~H. Szeto. 2024.
\newblock \href {https://doi.org/10.3389/fpsyt.2024.1336690} {Social perceptions and the stigmatization towards fifteen mental illnesses in {France}: A preliminary study on the role of vital force and burden}.
\newblock \emph{Frontiers in Psychiatry}, 15:1336690.

\bibitem[{Day et~al.(2007)Day, Edgren, and Eshleman}]{day2007measuring}
Emer~N. Day, Kara Edgren, and Amy Eshleman. 2007.
\newblock \href {https://doi.org/10.1111/j.1559-1816.2007.00255.x} {Measuring stigma toward mental illness: Development and application of the mental illness stigma scale}.
\newblock \emph{Journal of Applied Social Psychology}, 37(10):2191--2219.

\bibitem[{Fox et~al.(2018)Fox, Earnshaw, Taverna, and Vogt}]{fox2018conceptualizing}
Annie~B. Fox, Valerie~A. Earnshaw, Emily~C. Taverna, and Dawne Vogt. 2018.
\newblock \href {https://doi.org/10.1037/sah0000104} {Conceptualizing and measuring mental illness stigma: The mental illness stigma framework and critical review of measures}.
\newblock \emph{Stigma and Health}, 3(4):348--376.

\bibitem[{Giorgi et~al.(2024)Giorgi, Bellew, Habib, Sedoc, Smitterberg, Devoto, Himelein-Wachowiak, and Curtis}]{giorgi2024lived}
Salvatore Giorgi, Douglas Bellew, Daniel Roy~Sadek Habib, Jo{\~a}o Sedoc, Chase Smitterberg, Amanda Devoto, McKenzie Himelein-Wachowiak, and Brenda Curtis. 2024.
\newblock \href {https://doi.org/10.1609/icwsm.v18i1.31328} {Lived experience matters: Automatic detection of stigma toward people who use substances on social media}.
\newblock In \emph{Proceedings of the International AAAI Conference on Web and Social Media}, volume~18, pages 474--487, Buffalo, NY, USA. AAAI Press.

\bibitem[{Goulden et~al.(2011)Goulden, Corker, Evans-Lacko, Rose, Thornicroft, and Henderson}]{goulden2011}
Robert Goulden, Elizabeth Corker, Sara Evans-Lacko, Diana Rose, Graham Thornicroft, and Claire Henderson. 2011.
\newblock \href {https://doi.org/10.1186/1471-2458-11-796} {Newspaper coverage of mental illness in the {UK}, 1992--2008}.
\newblock \emph{BMC Public Health}, 11:796.

\bibitem[{Griffiths et~al.(2008)Griffiths, Christensen, and Jorm}]{griffiths2008predictors}
Kathleen~M. Griffiths, Helen Christensen, and Anthony~F. Jorm. 2008.
\newblock \href {https://doi.org/10.1186/1471-244X-8-25} {Predictors of depression stigma}.
\newblock \emph{BMC Psychiatry}, 8:25.

\bibitem[{Groot et~al.(2020)Groot, Rehm, Andrews, Hobern, Morgan, Green, Sweeney, and Blanchard}]{groot2020}
Christopher Groot, Imogen Rehm, Cal Andrews, Beth Hobern, Rikki Morgan, Hannah Green, Lisa Sweeney, and Michelle Blanchard. 2020.
\newblock Report on findings from the {Our Turn to Speak} survey: Understanding the impact of stigma and discrimination on people living with complex mental health issues.
\newblock Technical report, Anne Deveson Research Centre, SANE Australia, Melbourne, Australia.

\bibitem[{Jilka et~al.(2022)Jilka, Odoi, van Bilsen, Morris, Erturk, Cummins, Cella, and Wykes}]{jilka2022identifying}
Sagar Jilka, Clarissa~Mary Odoi, Janet van Bilsen, Daniel Morris, Sinan Erturk, Nicholas Cummins, Matteo Cella, and Til Wykes. 2022.
\newblock \href {https://doi.org/10.1038/s41537-021-00197-6} {Identifying schizophrenia stigma on {Twitter}: A proof of principle model using service user supervised machine learning}.
\newblock \emph{Schizophrenia}, 8(1).

\bibitem[{Kotov et~al.(2017)Kotov, Krueger, Watson, Achenbach, Althoff, Bagby, Brown, Carpenter, Caspi, Clark et~al.}]{kotov2017hierarchical}
Roman Kotov, Robert~F. Krueger, David Watson, Thomas~M. Achenbach, Robert~R. Althoff, R.~Michael Bagby, Timothy~A. Brown, William~T. Carpenter, Avshalom Caspi, Lee~Anna Clark, et~al. 2017.
\newblock \href {https://www.apa.org/pubs/journals/features/abn-abn0000258.pdf} {The hierarchical taxonomy of psychopathology ({HiTOP}): A dimensional alternative to traditional nosologies}.
\newblock \emph{Journal of Abnormal Psychology}, 126(4):454--477.

\bibitem[{Link and Phelan(2001)}]{link2001conceptualizing}
Bruce~G. Link and Jo~C. Phelan. 2001.
\newblock \href {https://doi.org/10.1146/annurev.soc.27.1.363} {Conceptualizing stigma}.
\newblock \emph{Annual Review of Sociology}, 27:363--385.

\bibitem[{Meng et~al.(2025)Meng, Chen, Li, Yang, Lee, Zhang, and Lee}]{meng2025stigma}
Han Meng, Yancan Chen, Yunan Li, Yitian Yang, Jungup Lee, Renwen Zhang, and Yi-Chieh Lee. 2025.
\newblock \href {https://doi.org/10.18653/v1/2025.acl-long.272} {What is stigma attributed to? {A} theory-grounded, expert-annotated interview corpus for demystifying mental-health stigma}.
\newblock In \emph{Proceedings of the 63rd Annual Meeting of the Association for Computational Linguistics (Volume 1: Long Papers)}, pages 5453--5490, Vienna, Austria. Association for Computational Linguistics.

\bibitem[{Mittal and De~Choudhury(2023)}]{Mittal2023}
Shravika Mittal and Munmun De~Choudhury. 2023.
\newblock \href {https://doi.org/10.1145/3544548.3580834} {Moral framing of mental health discourse and its relationship to stigma: A comparison of social media and news}.
\newblock In \emph{Proceedings of the 2023 CHI Conference on Human Factors in Computing Systems}, New York, NY, USA. Association for Computing Machinery.

\bibitem[{Moriceau et~al.(2022)Moriceau, Benamara, and Boumadane}]{moriceau2022automatic}
V{\'e}ronique Moriceau, Farah Benamara, and Abdelmoumene Boumadane. 2022.
\newblock \href {https://aclanthology.org/2022.lrec-1.25/} {Automatic detection of stigmatizing uses of psychiatric terms on {T}witter}.
\newblock In \emph{Proceedings of the Thirteenth Language Resources and Evaluation Conference}, pages 237--243, Marseille, France. European Language Resources Association.

\bibitem[{Oscar et~al.(2017)Oscar, Fox, Croucher, Wernick, Keune, and Hooker}]{oscar2017machine}
Nels Oscar, Pamela~A. Fox, Racheal Croucher, Riana Wernick, Jessica Keune, and Karen Hooker. 2017.
\newblock \href {https://doi.org/10.1093/geronb/gbx014} {Machine learning, sentiment analysis, and tweets: An examination of {Alzheimer's} disease stigma on {Twitter}}.
\newblock \emph{The Journals of Gerontology: Series B}, 72(5):742--751.

\bibitem[{Pavlova and Berkers(2022)}]{pavlova2022mental}
Alina Pavlova and Pauwke Berkers. 2022.
\newblock \href {https://doi.org/10.1080/10410236.2020.1862396} {``{Mental} health'' as defined by {Twitter}: Frames, emotions, stigma}.
\newblock \emph{Health Communication}, 37(5):637--647.

\bibitem[{Price(2022)}]{price2022}
Hazel Price. 2022.
\newblock \emph{The Language of Mental Illness: Corpus Linguistics and the Construction of Mental Illness in the Press}.
\newblock Cambridge University Press, Cambridge.

\bibitem[{{Reddit}(2024)}]{reddit2024privacy}
{Reddit}. 2024.
\newblock Reddit privacy policy.
\newblock \url{https://www.reddit.com/policies/privacy-policy}.
\newblock Accessed: 7 September 2024.

\bibitem[{Ritsher et~al.(2003)Ritsher, Otilingam, and Grajales}]{ritsher2003internalized}
Jennifer~Boyd Ritsher, Poorni~G. Otilingam, and Monica Grajales. 2003.
\newblock \href {https://doi.org/10.1016/j.psychres.2003.08.008} {Internalized stigma of mental illness: Psychometric properties of a new measure}.
\newblock \emph{Psychiatry Research}, 121(1):31--49.

\bibitem[{Sartorius(2007)}]{sartorius2007stigma}
Norman Sartorius. 2007.
\newblock \href {https://doi.org/10.1016/S0140-6736(07)61245-8} {Stigma and mental health}.
\newblock \emph{The Lancet}, 370(9590):810--811.

\bibitem[{Schomerus et~al.(2011)Schomerus, Lucht, Holzinger, Matschinger, Carta, and Angermeyer}]{schomerus2011stigma}
Georg Schomerus, Michael Lucht, Anita Holzinger, Herbert Matschinger, Mauro~G. Carta, and Matthias~C. Angermeyer. 2011.
\newblock \href {https://doi.org/10.1093/alcalc/agq089} {The stigma of alcohol dependence compared with other mental disorders: A review of population studies}.
\newblock \emph{Alcohol and Alcoholism}, 46(2):105--112.

\bibitem[{Straton et~al.(2020)Straton, Jang, and Ng}]{straton2020stigma}
Nadiya Straton, Hyeju Jang, and Raymond Ng. 2020.
\newblock \href {https://aclanthology.org/2020.lrec-1.148/} {Stigma annotation scheme and stigmatized language detection in health-care discussions on social media}.
\newblock In \emph{Proceedings of the Twelfth Language Resources and Evaluation Conference}, pages 1178--1190, Marseille, France. European Language Resources Association.

\bibitem[{Taylor and Dear(1981)}]{taylor1981scaling}
S.~Martin Taylor and Michael~J. Dear. 1981.
\newblock \href {https://doi.org/10.1093/schbul/7.2.225} {Scaling community attitudes toward the mentally ill}.
\newblock \emph{Schizophrenia Bulletin}, 7(2):225--240.

\bibitem[{Thornicroft et~al.(2022)Thornicroft, Sunkel, Alikhon~Aliev, Baker, Brohan, el~Chammay, Davies, Demissie, Duncan, Fekadu et~al.}]{thornicroft2022lancet}
Graham Thornicroft, Charlene Sunkel, Akmal Alikhon~Aliev, Sue Baker, Elaine Brohan, Rabih el~Chammay, Kelly Davies, Mekdes Demissie, Joshua Duncan, Wubalem Fekadu, et~al. 2022.
\newblock \href {https://doi.org/10.1016/S0140-6736(22)01470-2} {The {Lancet Commission} on ending stigma and discrimination in mental health}.
\newblock \emph{The Lancet}, 400(10361):1438--1480.

\bibitem[{Wanniarachchi et~al.(2023)Wanniarachchi, Scogings, Susnjak, and Mathrani}]{wanniarachchi2023hate}
Vajisha~Udayangi Wanniarachchi, Chris Scogings, Teo Susnjak, and Anuradha Mathrani. 2023.
\newblock \href {https://ajis.aaisnet.org/index.php/ajis/article/view/3929} {Hate speech patterns in social media: A methodological framework and fat stigma investigation incorporating sentiment analysis, topic modelling and discourse analysis}.
\newblock \emph{Australasian Journal of Information Systems}, 27.

\end{thebibliography}

\appendix

\section{Corpus preprocessing} 
\label{sec:appendix_preprocessing}

This appendix reports preprocessing procedures and descriptive counts for the two online communication corpora used in the study: the News on the Web (NOW) corpus and the Reddit corpus.

\subsection{Targets}

Given condition-specific variation in mental health stigma, our final target set comprised six mental health conditions: ADHD, autism spectrum disorder, borderline personality disorder, depression, alcoholism, and schizophrenia. These were selected to balance theoretical coverage, corpus frequency, and expected variation in stigma. We began by compiling candidates from major HiTOP spectra \cite{kotov2017hierarchical}---internalizing, neurodevelopmental, psychotic, personality, and substance-related---and, in consultation with domain experts, narrowed the list, consolidating substance-related terms under ``alcoholism.'' For each condition, we generated orthographic and acronymic variants, ran pilot searches and manual inspections in the NOW corpus, normalized hyphens, slashes, and case, and excluded spurious hits such as ``BPD'' referring to ``barrels per day'' in petroleum reports. Manual inspection of 300 articles per target confirmed that these variants improved article relevance, meaning that articles mentioned the condition in a more-than-incidental way. Table~\ref{tab:search_lexicon} reports the search lexicon for each mental health condition.


\begin{table}[h]
\centering
\small
\setlength{\tabcolsep}{4pt}
\renewcommand{\arraystretch}{1.08}
\begin{tabularx}{\columnwidth}{@{}lX@{}}
\toprule
\textbf{Condition} & \textbf{Search lexicon} \\
\midrule
ADHD & ADHD; attention deficit hyperactivity disorder \\
Alcoholism & alcoholism; alcohol abuse; alcohol dependence; alcohol use disorder \\
Autism & autism; autistic; autism spectrum disorder; ASD \\
BPD & borderline personality disorder; borderline PD \\
Depression & depression; major depressive disorder; clinical depression \\
Schizophrenia & schizophrenia; schizophrenic \\
\bottomrule
\end{tabularx}
\caption{Surface variants used to retrieve candidate texts for each target mental health condition. Variants were grouped under their associated target condition before sampling and annotation.}
\label{tab:search_lexicon}
\end{table}

\subsection{News on the Web}

The News on the Web (NOW) corpus is a compilation of online news articles drawn from over 7,000 outlets across 20 countries, spanning 2010–2025. For the present investigation, we extracted all items containing any surface variants from our six target mental health terms, thereby constructing six dedicated sub-corpora.  In total, the NOW mental illness conditions sub-corpus comprises 166,912 news articles. We focused on articles from US newspapers. 

Pre-processing steps entailed maximizing the reliability of the NOW‐corpus material used for stigma annotation. First, within the depression subset we discarded 826 articles that contained the bigram "economic depression", thereby removing economic, rather than clinical, references to “depression.” Next, any article that mentioned more than one diagnostic target (e.g., both ADHD and autism) was excluded so that subsequent judgments could be unambiguously attributed to a single disorder. For borderline personality disorder, this yielded under the 500 random sample threshold (427 articles). We also eliminated every record bearing the "TOOLONG" flag, which marks texts that were truncated during corpus compilation and, hence, provide incomplete linguistic evidence. For the texts that were  retained, we extracted only the first paragraph in which the target term appears and the two paragraphs that immediately follow it, yielding focused, context-rich passages for human coding (and as for news articles, important content comes first). Finally, the material was lightly cleaned: superfluous whitespace around punctuation, parentheses, quotation marks, and apostrophes was normalized, and residual HTML markers were stripped. Table~\ref{tab:unique_articles} reports the number of unique articles in which each target appears.

\begin{table}[h]
  \centering
  \footnotesize
  \begin{tabular}{l r}
    \toprule
    \textbf{Target}    & \textbf{Articles} \\
    \midrule
    ADHD               & 8,316   \\
    alcoholism         & 12,934  \\
    autism             & 26,405  \\
    borderline personality disorder     & 976      \\
    depression         & 106,193 \\
    schizophrenia      & 12,088  \\
    \bottomrule
  \end{tabular}
  \caption{Articles per target in the NOW subcorpora}
  \label{tab:unique_articles}
\end{table}


Detailed counts of surface-variant occurrences for each mental health condition and its article counts are provided in Table~\ref{tab:variant_matches}.


\begin{table}[h]
  \centering
  \footnotesize
  \begin{tabular}{l r}
    \toprule
    \textbf{Variant}                                  & \textbf{Articles} \\
    \midrule
    ADHD                                              & 7,811  \\
    attention deficit hyperactivity disorder          & 1,455   \\
    \midrule
    alcoholism                                        & 8,674  \\
    alcohol abuse                                     & 4,305   \\
    alcohol dependence                                & 463      \\
    alcohol use disorder                              & 837   \\
    \midrule
    autism                                            & 21,278  \\
    autistic                                          & 8,251  \\
    autism spectrum disorder                          & 2,703   \\
    ASD                                               & 1,888  \\
    \midrule
    borderline personality disorder                    & 972   \\
    borderline PD                                     & 12       \\
    \midrule
    depression                                        & 105,600 \\
    major depressive disorder                         & 1,712   \\
    clinical depression                               & 1,307   \\
    \midrule
    schizophrenia                                     & 9,945  \\
    schizophrenic                                     & 2,683   \\
    \bottomrule
  \end{tabular}
  \caption{Articles containing at least one instance of the surface variant in its associated NOW subcorpus}
  \label{tab:variant_matches}
\end{table}

\begin{table}[h]
\centering
\small
\renewcommand{\arraystretch}{1.2}
\begin{tabular}{lrrrrrr}
\toprule
\textbf{Target} & \textbf{Median (whole)} & \textbf{Median (sampled)} \\
\midrule
ADHD           & 1,265  & 1,284  \\
Alcoholism     & 1,142  & 1,199  \\
Autism         & 1,050  & 976  \\
Borderline PD  & 1,233  & 1,221  \\
Depression     & 1,145  & 1,071 \\
Schizophrenia  & 1,084  & 1,030 \\
\bottomrule
\end{tabular}
\caption{Article length descriptives (token count) for all articles in the full and randomly sampled NOW corpora per target.}
\label{tab:fullcorpus_article_lengths}
\end{table}


\FloatBarrier
\subsection{Reddit}

Reddit is a forum-style social media platform organized around topic-specific communities called subreddits. It is a widely used source for research on online mental health discourse due to its large volume of user-generated content. Our dataset was collected via Pushshift.io and covered the years of 2012–2022. To construct the corpus, we extracted posts of at least 30 words containing any of the six target mental health terms, excluding posts that mentioned more than one of these terms to ensure each post was associated with a single condition. Unlike the NOW-corpus texts, Reddit posts were retained in full rather than truncated. In total, the Reddit mental health sub-corpus comprises 1,303,437 posts. Post counts by target condition are reported in Table~\ref{tab:reddit_posts}.

\begin{table}[th]
  \centering
  \begin{tabular}{l r}
    \toprule
    \textbf{Target}    & \textbf{Posts} \\
    \midrule
    ADHD               & 219,715   \\
    Alcoholism         & 45,712  \\
    Autism             & 122,467  \\
    Borderline PD     & 91,714      \\
    Depression         & 792,623 \\
    Schizophrenia      & 31,206  \\
    \bottomrule
  \end{tabular}
  \caption{Posts per target in the Reddit subcorpora}
  \label{tab:reddit_posts}
\end{table}

\begin{table}[ht]
\centering
\label{tab:text-length}
\begin{tabular}{lrr}
\toprule
\textbf{Target} & \textbf{Median} & \textbf{Q3} \\
\midrule
ADHD          &   992 & 1,792 \\
Alcoholism    & 1,580 & 2,996 \\
Autism        & 1,142 & 2,181 \\
Borderline PD           & 1,225 & 2,328 \\
Depression    & 1,343 & 2,539 \\
Schizophrenia & 1,264 & 2,675 \\
\bottomrule
\end{tabular}
\caption{Post length descriptives for Reddits subcorpora per target.}
\end{table}

\twocolumn

\section{Stigma taxonomy} 
\label{appendix:appendix_stigmataxonomy}

The taxonomy combines source frameworks and measures at different levels. Stigma mode follows the Mental Illness Stigma Framework \cite{fox2018conceptualizing}; stigma domains and public stigma sub-factors follow the Lancet Commission taxonomy \cite{thornicroft2022lancet}; and the public-stigma components were adapted from established stigma measures, including the Attribution Questionnaire \cite{corrigan2003attribution}, the Depression Stigma Scale \cite{griffiths2008predictors}, the Community Attitudes toward the Mentally Ill scale \cite{taylor1981scaling}, the Internalized Stigma of Mental Illness scale \cite{ritsher2003internalized}, and the Mental Illness Stigma Scale \cite{day2007measuring}. The prototype exemplar sources are summarised in Appendix~\ref{appendix:prototype_exemplars}.

\onecolumn

\small
\renewcommand{\arraystretch}{1.2}

\begin{longtable}{p{0.10\textwidth}p{0.22\textwidth}p{0.40\textwidth}p{0.17\textwidth}}
\toprule
\textbf{Level} & \textbf{Type} & \textbf{Definition and Example} & \textbf{Source} \\
\midrule
\endfirsthead

\toprule
\textbf{Level} & \textbf{Type} & \textbf{Definition and Example} & \textbf{Source} \\
\midrule
\endhead

\multicolumn{4}{l}{\textbf{Level 2: Stigma Mode}} \\
\midrule

& \textbf{Expressed/Enacted} 
& Expression of stigma or description of its enactment by a stigmatizer. \textit{Example:} ``Police restrained the man using force after noting he had a history of psychiatric admissions.'' 
& \cite{fox2018conceptualizing} \\

& \textbf{Perceived/Experienced} 
& Stigma is perceived or its expression experienced by people with lived experience. \textit{Example:} ``I was treated like I was dangerous,'' said Mark, who lives with bipolar disorder. 
& \cite{fox2018conceptualizing} \\

\midrule
\multicolumn{4}{l}{\textbf{Level 3: Stigma Domain}} \\
\midrule

& \textbf{Public} 
& Negative and misinformed stereotyped beliefs, prejudicial emotional responses, or discriminatory behaviors. \textit{Example:} ``Dr White said that, `Once people develop mental illness, they will never recover completely.' '' 
& \cite{thornicroft2022lancet} \\

& \textbf{Structural} 
& Systems, policies, or laws that disempower, discriminate, or limit opportunities for people with lived experience of mental illness. \textit{Example:} ``The report found that mental health patients waited twice as long for emergency care.'' 
& \cite{thornicroft2022lancet} \\

& \textbf{Associative} 
& Stigma is extended to family, carers, supporters, health workers, or others around people with lived experience of mental illness. \textit{Example:} ``Mental illness is often concealed by families for fear of judgment.'' 
& \cite{thornicroft2022lancet} \\

& \textbf{Self} 
& Internalization of negative public stereotypes. \textit{Example:} ``For a long time, I thought my mental illness made me a burden to everyone around me.'' 
& \cite{thornicroft2022lancet} \\

\midrule
\multicolumn{4}{l}{\textbf{Level 4: Public Stigma Sub-factors}} \\
\midrule

& \textbf{Cognitive (Stereotypes)} 
& Stereotypes are socially shared and generalized negative beliefs about mental illness and people living with it; they represent the cognitive component of stigma. \textit{Example:} ``He could snap out of it if he just tried.'' 
& \cite{thornicroft2022lancet} \\

& \textbf{Affective (Emotions)} 
& Prejudicial affective responses are emotional reactions that arise from stereotypes about mental illness and people living with it; they represent the affective component of stigma. \textit{Example:} ``Many feel uneasy around people who are mentally ill.'' 
& \cite{thornicroft2022lancet} \\

& \textbf{Behavioral (Discrimination)} 
& Discriminatory behaviors are the actions that result from stereotypes and prejudicial emotions about mental illness and people living with it; they constitute the behavioral component of stigma. \textit{Example:} ``The employer admitted he wouldn't hire someone with a history of mental illness.'' 
& \cite{thornicroft2022lancet} \\

\midrule
\multicolumn{4}{l}{\textbf{Level 5: Public Stigma Components}} \\
\midrule

& \textbf{Cognitive -- Dangerousness} 
& A belief that someone living with mental illness is dangerous. \textit{Example:} ``I would feel that Harry would pose a threat to others.'' 
& \cite{corrigan2003attribution} \\

& \textbf{Cognitive -- Personal responsibility and blame} 
& A belief that someone living with mental illness is responsible and/or to blame for their experience. \textit{Example:} ``I would think that it was Harry's fault that he is in the present condition.'' 
& \cite{corrigan2003attribution} \\

& \textbf{Cognitive -- Uncontrollable} 
& A belief that someone is not in control of their experience of mental illness and the behavior that may arise from it. \textit{Example:} ``I would feel that Harry is unpredictable.'' 
& \cite{corrigan2003attribution} \\

& \textbf{Cognitive -- Controllable} 
& A belief that someone is in control of their experience of mental illness and the behavior that may arise from it. \textit{Example:} ``People with depression could snap out of it if they wanted.'' 
& \cite{griffiths2008predictors} \\

& \textbf{Cognitive -- Moral or character flaw} 
& A belief that a person's mental illness is a reflection of a moral or character flaw. \textit{Example:} ``Depression is a sign of personal weakness.'' 
& \cite{griffiths2008predictors} \\

& \textbf{Cognitive -- Pity/condescension} 
& A belief that someone living with mental illness is inferior or less capable. \textit{Example:} ``I would feel sympathy for Harry.'' 
& \cite{corrigan2003attribution} \\ 

& \textbf{Cognitive -- Trivialization or minimization} 
& A belief that someone's mental illness is not a real, legitimate, sufficiently severe, or impactful condition. \textit{Example:} ``Depression is not a real mental illness.'' 
& \cite{griffiths2008predictors} \\

& \textbf{Cognitive -- Immutability} 
& A belief that a person cannot recover from their experience of mental illness. \textit{Example:} ``People with mental illness cannot live a good, rewarding life.'' 
& \cite{ritsher2003internalized,day2007measuring} \\

& \textbf{Cognitive -- Segregation endorsement} 
& A belief that people with mental illness should be physically or socially separated from the community. \textit{Example:} ``I would think that Harry should be kept away from others.'' 
& \cite{corrigan2003attribution} \\

& \textbf{Cognitive -- Coercion} 
& A belief that people with mental illness must be controlled or forced into treatment because they lack insight or self-control. \textit{Example:} ``I would force Harry to take his medication.'' 
& \cite{corrigan2003attribution} \\

& \textbf{Cognitive -- Contamination} 
& A belief that mental illness is contagious or morally polluting, or that association with someone with mental illness may damage one's status or well-being. \textit{Example:} ``It is best to avoid people with depression so you don't become depressed yourself.'' 
& \cite{griffiths2008predictors} \\

& \textbf{Cognitive -- Incompetence} 
& A belief that people with mental illness are incapable or unreliable in daily or occupational functioning. \textit{Example:} ``People with a mental illness should not be given any responsibilities.'' 
& \cite{taylor1981scaling} \\

& \textbf{Affective -- Fear} 
& An emotional response of fear in response to people living with mental illness. \textit{Example:} ``I would feel afraid of Harry.'' 
& \cite{corrigan2003attribution} \\

& \textbf{Affective -- Anger, irritability, or frustration} 
& An emotional response of anger, irritability, or frustration in response to people living with mental illness. \textit{Example:} ``I would feel irritated by Harry.'' 
& \cite{corrigan2003attribution} \\

& \textbf{Affective -- Disgust} 
& An emotional response of revulsion or moral contamination toward people with mental illness. \textit{Example:} ``I would feel sickened by them.'' 
& Theory-derived \\

& \textbf{Affective -- Pity} 
& A compassionate but patronising emotional response towards people with mental illness. \textit{Example:} ``I would feel pity for Harry.'' 
& \cite{corrigan2003attribution} \\

& \textbf{Affective -- Ambivalence} 
& A mixed or conflicted emotional response involving simultaneous positive and negative feelings toward people with mental illness. \textit{Example:} ``I would feel both compassion and discomfort toward them.'' 
& Theory-derived \\

& \textbf{Behavioral -- Social distancing} 
& Behavioral discrimination involving desire for, or actual limitation of, the degree to which someone with mental illness is included in one's life across numerous contexts. \textit{Example:} ``I would stay away from Harry.'' 
& \cite{corrigan2003attribution} \\

& \textbf{Behavioral -- Bullying/mocking} 
& Behavioral discrimination involving actual verbal or physical abuse of a person with a mental health condition. \textit{Example:} ``People mocked him for being mentally ill.'' 
& Theory-derived \\

\bottomrule
\caption{Definitions and illustrative examples of stigma expressions across hierarchical levels. Citations indicate the source framework, scale, or measure from which each category or illustrative item was adapted.}
\label{tab:appendix_stigma_taxonomy}
\end{longtable}

\normalsize

\newpage
\twocolumn
\section{Prototype exemplar sources}
\label{appendix:prototype_exemplars}

The prototype exemplar dataset contains stigma-present and stigma-absent usages for each annotation category. Prototype usages were mapped to the taxonomy from examples based on validated stigma scale items and were expanded where necessary to provide at least ten stigma-present and ten stigma-absent usages per category. Expansion primarily involved creating close variants of validated scale items and generating stigma-absent usages by negating or reframing stigma-present usages. Table~\ref{tab:prototype_sources} summarizes the source measures used to construct the prototype exemplar set. The full prototype dataset contains stigma-present and stigma-absent examples for each annotation category and is provided at this link: \url{https://github.com/jemimakang/mh_stigma/blob/main/data/prototype_usages_stigma_types.csv}.
\onecolumn

\begin{table*}[h]
\centering
\small
\renewcommand{\arraystretch}{1.2}
\begin{tabularx}{\textwidth}{p{0.28\textwidth}p{0.42\textwidth}X}
\toprule
\textbf{Source measure or framework} & \textbf{Categories informed} & \textbf{Citation} \\
\midrule
Attribution Questionnaire & Dangerousness; personal responsibility and blame; uncontrollable; pity/condescension; segregation endorsement; coercion; fear; anger/irritation; pity; social distancing & \cite{corrigan2003attribution} \\
Depression Stigma Scale & Perceived/experienced stigma; controllability; moral or character flaw; trivialization/minimization; contamination; selected dangerousness and blame prototypes & \cite{griffiths2008predictors} \\
Community Attitudes toward the Mentally Ill scale & Structural stigma; public stigma prototypes; dangerousness; moral or character flaw; trivialization/minimization; segregation; coercion; incompetence; fear & \cite{taylor1981scaling} \\
Internalized Stigma of Mental Illness scale & Self-stigma; internalized stereotypes; affective responses; behavioral withdrawal; immutability; selected incompetence and social-withdrawal prototypes & \cite{ritsher2003internalized} \\
Mental Illness Stigma Scale & Immutability and other general mental-illness stigma components where applicable & \cite{day2007measuring} \\
Theory-derived definitions & Expressed/enacted stigma; public, structural, and associative stigma; broad cognitive, affective, and behavioral sub-factors; disgust; ambivalence & Present taxonomy \\
\bottomrule
\end{tabularx}
\caption{Source measures and frameworks used to construct the stigma-present and stigma-absent prototype exemplar set.}
\label{tab:prototype_sources}
\end{table*}

\twocolumn

\section{Annotation protocol and guideline revisions}
\label{app:annotation_procedure}

This appendix provides additional details on the annotation codebook and iterative annotation process summarized in Section~\ref{sec:annotation}. The codebook contained category definitions, decision rules, and examples for the binary stigma-presence task and the fine-grained stigma taxonomy. The complete codebook is provided here: \url{https://github.com/jemimakang/mh_stigma/blob/main/annotation/annotation_instructions.pdf}.


\paragraph{Operational definition of stigma.} 

For the binary stigma task, stigma was annotated as present when a text contained at least one of the following, directed at a person or group with a mental health condition, or those associated with them:
(i) a negative trait attribution, such as being dangerous, weak, incompetent, or unpredictable; 
(ii) a negative emotional reaction, such as fear, disgust, contempt, or ridicule; 
(iii) negative treatment or discrimination, such as calls to isolate, punish, or deny rights or services, including discriminatory laws, rules, or institutional practices; 
(iv) derogatory or demeaning language, such as slurs or insults; or 
(v) intrinsically stigmatizing uses of a mental health condition term, such as metaphorical or pathologizing uses outside a clinical context.

Texts were annotated as stigma-absent when the negative attribution, reaction, or treatment was not based on or related to mental health status; when the text merely discussed, criticized, or reflected on stigma without displaying it; or when the negative description was situation-specific and did not imply a generalized stigmatizing attitude toward people with mental illness. Texts that described an incident of stigma perpetrated by someone else were coded as stigma-present only when the passage itself displayed stigmatizing attribution, treatment, or language.

\paragraph{Annotation rounds.}

In Round 1, annotators independently annotated news and Reddit texts using the initial codebook. They judged whether stigma was present and, where relevant, selected applicable fine-grained stigma categories. Majority judgments were used to identify candidate stigma-present texts, which were then reviewed and adjudicated by two researchers. The adjudication process selected the final label where there was agreement and consulted annotators in ambiguous cases or cases involving multiple plausible codes. Textual evidence spans were also identified for stigma-present cases.

Because random sampling in the first round yielded a large proportion (85\%) of stigma-absent examples, later rounds used more targeted strategies to enrich the benchmark with stigma-present and boundary cases. In Round 2, randomly sampled candidate texts were pre-filtered using an LLM prompt containing stigma definitions and examples. This yielded 350 candidate samples predicted by GPT-5.2 Pro to contain stigma.\footnote{Inference details for GPT-5.2 Pro are reported in Appendix~\ref{app:model_details}.} After manual screening to remove irrelevant or clearly non-stigmatizing cases, and after shortening overly long posts to retain only the necessary context for annotation, 231 texts were retained for annotation. Some borderline or likely stigma-absent texts were deliberately retained as validity checks.


Round 3 introduced the explicit operational decision rule described above to improve consistency across annotators. In particular, annotators were instructed to distinguish texts that displayed or described stigma from texts that merely discussed, critiqued, or reflected on stigma as a topic. Texts were labeled stigma-present only when the negative attribution, emotional reaction, treatment, or demeaning language was tied to mental health status rather than to unrelated behavior or situational context.


After Round 3, we examined annotator disagreements and found that a major source of inconsistency concerned the distinction between stigma modes, especially \textit{expressed/enacted} versus \textit{perceived/experienced} stigma. We therefore revised the guidelines to make the mode decision rule more explicit. The revised rule defined \textit{expressed/enacted} stigma as cases in which a stigmatizing source directly expressed stigma or in which the text described stigma being enacted by a person, institution, policy, system, or media text. It defined \textit{perceived/experienced} stigma as cases in which a person with lived or living experience described being stereotyped, devalued, discriminated against, feared, excluded, mocked, or otherwise treated negatively because of their condition.


In Round 4, we further enhanced the benchmark by sampling texts on which automatic systems showed the greatest disagreement. Disagreement was defined across predictions from Claude Sonnet 4.6, Gemini 3 Flash, and GPT-5.5. Candidate texts were then manually filtered before annotation. Before this round, annotators completed a training exercise focused on threshold cases for \textit{stigma mode}, using the revised mode decision rule. They then annotated the selected texts using the updated guidelines. 

\paragraph{Operational rules for stigma mode.}
Stigma mode captured whether stigma was \textit{expressed/enacted} by a stigmatizing source, \textit{perceived/experienced} by a person with lived or living experience of mental illness, or \textit{both}. Expressed/enacted stigma was annotated when a text expressed stigma directly or described stigma being enacted by a person, institution, policy, system, or media text. Perceived/experienced stigma was annotated when a text described stigma from the perspective of a person with lived or living experience of mental illness, including reports of being stereotyped, devalued, discriminated against, feared, excluded, mocked, or otherwise treated negatively because of their condition. When self-stigma was annotated, stigma mode was coded as both expressed/enacted and perceived/experienced because self-stigma involves a person with lived experience internalizing stigma while also reproducing or expressing a negative stereotype about themselves.


\subsection{Annotation platform interface}
Annotation was conducted using Label Studio (see Figure 2 for an example of the interface). Each task presented annotators with a target text alongside its associated mental health term. The interface was structured hierarchically, whereby each response revealed only the questions relevant to the annotator's preceding choice. A free-text field was available throughout all levels to capture any potential observations or comments from the annotators. 

Annotators first made a binary judgment of whether stigma was present in the text. They then rated their confidence in this judgment (0--10). Text judged as containing no stigma required no further annotation. Where stigma was identified, annotators proceeded through the following levels of categorization.

\subsubsection{Mode} Annotators indicated whether stigma was expressed/enacted, perceived/experienced, or both.

\subsubsection{Domain} Annotators selected one or more domains from four categories: public, structural, associative, and self-stigma.

\subsubsection{Sub-factor and component (public domain only)} Where public stigma was selected, annotators identified one or more sub-factors (cognitive, affective, or behavioral) and then selected the specific components within each. The cognitive sub-factor comprised 12 components (e.g., dangerousness, blame, uncontrollable, moral flaw); the affective sub-factor comprised five components (e.g., fear, disgust, pity); and the behavioral sub-factor included two components: social distancing and bullying. For each sub-factor, an ``Other'' option was included alongside the predefined components to capture any not represented in the lists. Selecting this option revealed a free-text field in which annotators could enter one or more component labels of their own.

Annotators could highlight relevant text spans using color-coded labels corresponding to their selections: component-level labels for public stigma and domain-level labels for structural, associative, and self-stigma.

\onecolumn
\begin{figure}[H]
    \centering
    \includegraphics[width=0.85\linewidth]{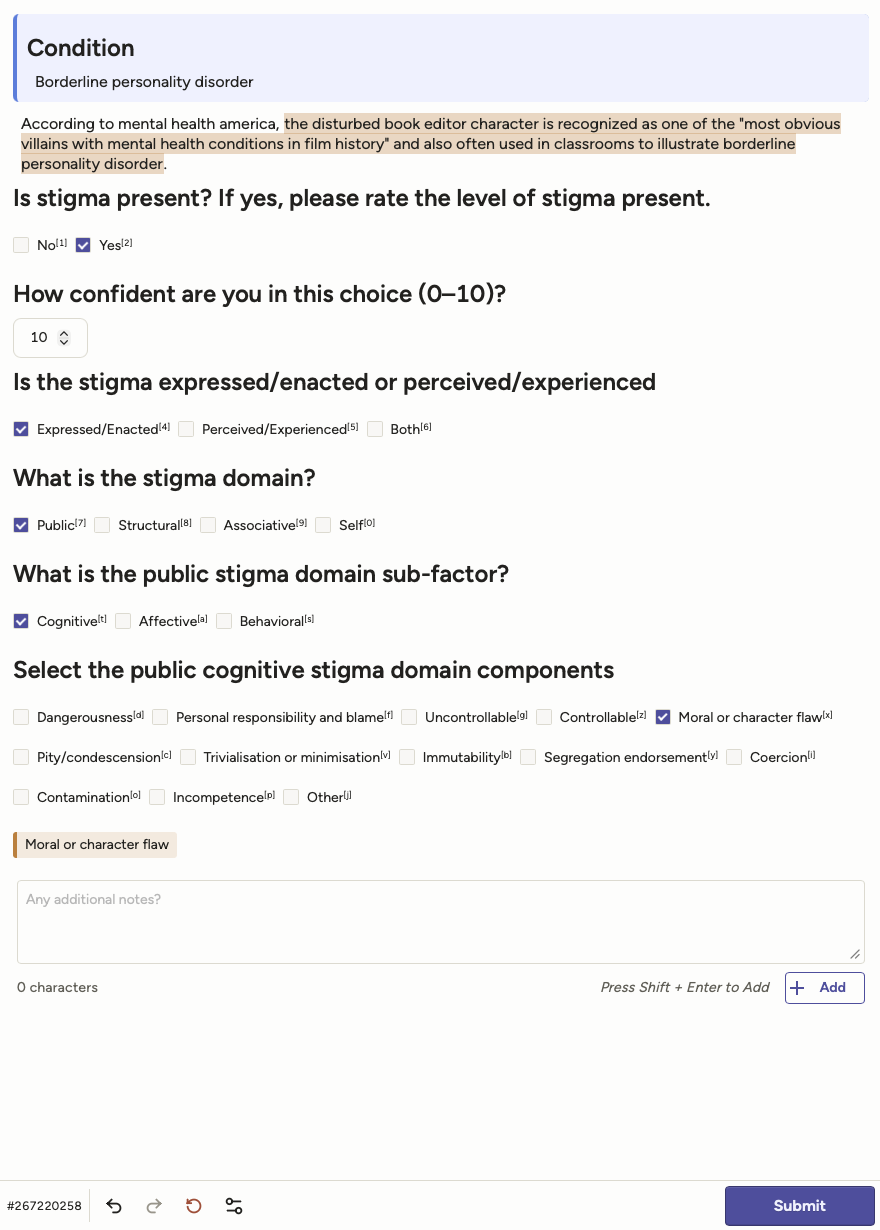}
    \caption{Annotation interface example.}
    \label{fig:Annotation platform example}
\end{figure}

\newpage

\section{Stigma detection and classification prompts}

\subsection{Prompts for binary classification of stigma presence}
\label{app:binary_prompts}

\subsection{``Definition'' prompt}

\begin{prompt}

According to psychologists "Stigma exists when elements of labeling, stereotyping, separation, status loss, and discrimination occur together".\\

Using this definition, does the following text contain stigma?

\end{prompt}

\subsection{``Examples'' prompt}


\begin{prompt}

Here are some examples of sentences that contain stigma according to Psychologists:\\

"My roommate keeps calling me ‘the anxious one’ to their friends. Like that is all I am."

"The schizophrenic was taken into custody."

"Police are concerned the suspect may have mental illness, which would explain the violent behavior."

"Someone in my class at uni said people with schizophrenia are all violent."

"Society needs better protection from the insane."

"I just wish my depressed boyfriend could be normal like everyone else."

"Employers have the right to know if someone is mentally unstable before hiring them."

"The landlord knocked back my application. He found out that I have a mental illness."

"Experts decided patients should not be consulted."

"My psychiatrist changed my medication without asking me what I thought. I feel so helpless."\\

Using these examples, does the following text contain stigma?

\end{prompt}

\subsection{``Definition+examples'' prompt}

\begin{prompt}
According to psychologists "Stigma exists when elements of labeling, stereotyping, separation,
status loss, and discrimination occur together".\\

Here are some examples of sentences that contain stigma according to Psychologists:\\

"My roommate keeps calling me ‘the anxious one’ to their friends. Like that is all I am."

"The schizophrenic was taken into custody."

"Police are concerned the suspect may have mental illness, which would explain the violent behavior."

"Someone in my class at uni said people with schizophrenia are all violent."

"Society needs better protection from the insane."

"I just wish my depressed boyfriend could be normal like everyone else."

"Employers have the right to know if someone is mentally unstable before hiring them."

"The landlord knocked back my application. He found out that I have a mental illness."

"Experts decided patients should not be consulted."

"My psychiatrist changed my medication without asking me what I thought. I feel so helpless."\\

Using this definition and these examples, does the following text contain stigma?
\end{prompt}

\subsection{``Definition+examples+rule'' prompt}

\begin{prompt}

According to psychologists "Stigma exists when elements of labeling, stereotyping, separation,
status loss, and discrimination occur together".\\

Here are some examples of sentences that contain stigma according to Psychologists:\\

"My roommate keeps calling me ‘the anxious one’ to their friends. Like that is all I am."

"The schizophrenic was taken into custody."

"Police are concerned the suspect may have mental illness, which would explain the violent behavior."

"Someone in my class at uni said people with schizophrenia are all violent."

"Society needs better protection from the insane."

"I just wish my depressed boyfriend could be normal like everyone else."

"Employers have the right to know if someone is mentally unstable before hiring them."

"The landlord knocked back my application. He found out that I have a mental illness."

"Experts decided patients should not be consulted."

"My psychiatrist changed my medication without asking me what I thought. I feel so helpless."\\

\#\# Stigma Present

Only annotate 'yes' if the text contains at least one of the following directed at a person or persons with a mental health condition (or those associated with them):\\

- **Negative trait attribution** (e.g., dangerous, weak, incompetent, unpredictable)

- **Negative emotional reaction** (e.g., fear, disgust, contempt, ridicule)

- **Negative treatment or discrimination** (e.g., calls to isolate, punish, or deny rights/services, including discriminatory laws, rules, or institutional practices)

- **Derogatory or demeaning language** (e.g., slurs, insults)\\

Annotate 'no' if:\\

- The negative attribution, reaction or treatment is not based on or related to the person’s mental health condition (e.g., anger is expressed towards a person with depression because they committed a crime that is unrelated to their depression).

- The negative attribution, reaction or treatment is mentioned in text rather than endorsed or expressed (e.g., a journalist writes about negative attitudes towards schizophrenia without manifesting it herself; a person criticizing or commenting on stigma). *Note:* text that simply describes an incident of stigma perpetrated by someone else should be annotated YES: the key issue is whether the text displays stigma or comments/reflects on it. 

\end{prompt}

\twocolumn
\section{Model and inference details}
\label{app:model_details}

Table~\ref{tab:model_details} reports the model identifiers and inference settings used in the LLM experiments. All model runs were conducted in May 2026. Unless otherwise stated, decoding and reasoning parameters were not manually tuned; models were run with provider or software-wrapper defaults. Model parameter counts are not publicly disclosed for the proprietary models evaluated here. We therefore identify models by their public names, API identifiers, and inference settings rather than by parameter size. The same prompt templates were used across models within each experimental condition.

\begin{table*}[h]
\centering
\small
\setlength{\tabcolsep}{5pt}
\renewcommand{\arraystretch}{1.18}
\begin{tabularx}{\textwidth}{@{}p{0.08\textwidth}p{0.24\textwidth}X@{}}
\toprule
\textbf{Provider} & \textbf{Model} & \textbf{Inference settings} \\
\midrule

Anthropic 
& Claude Sonnet 4.6 \texttt{claude-sonnet-4-6}
& Temperature was set to 0. Adaptive/extended thinking was disabled. Safety settings used provider defaults. \\

\midrule

OpenAI 
& GPT-5.5 \texttt{gpt-5.5}
& Temperature was  was set to 0. Reasoning effort was not explicitly overridden and therefore used the documented GPT-5.5 default of \texttt{medium}. Reasoning summaries were not requested. Safety settings used provider defaults. \\

\midrule

Google 
& Gemini 3 Flash \texttt{gemini-3-flash-preview}
& Temperature was set to 0. Dynamic thinking was used at default settings (\texttt{high}). Safety settings used provider defaults. \\

\midrule

DeepSeek 
& DeepSeek V4 Flash \texttt{deepseek-v4-flash}
& The model was used in thinking mode, where temperature and top-$p$ are not supported. Safety settings used provider defaults. \\
\midrule

OpenAI 
& GPT-5.2 Pro \texttt{gpt-5.2-pro-2025-12-11}
& Used only for LLM-assisted candidate filtering during benchmark construction, not for the main evaluation. Temperature = 0. Reasoning effort was not explicitly overridden in our filtering run. Safety settings used provider defaults. \\

\bottomrule
\end{tabularx}

\caption{Model identifiers and inference settings used in the LLM experiments. All model runs were conducted in May 2026. Unless otherwise stated, decoding, reasoning, and safety parameters were not manually tuned; where provider defaults or wrapper behavior were not fully exposed, we report the settings as unset rather than inferred.}
\label{tab:model_details}
\end{table*}

\newpage

\section{Benchmark composition: Label distributions}
\label{app:label_distribution}

This appendix provides expanded descriptive statistics for the benchmark. As in the main text, these distributions describe benchmark composition rather than stigma prevalence in the underlying corpora. Because stigma-present texts could receive multiple labels, counts for domains, subfactors, and components should be interpreted as text-label assignments rather than mutually exclusive classes.

Table~\ref{tab:benchmark_composition_expanded} provides the full benchmark composition, including source, condition, stigma mode, higher-level taxonomy labels, and fine-grained public-stigma components.

\begin{table*}[p]
\centering
\footnotesize
\setlength{\tabcolsep}{4pt}
\renewcommand{\arraystretch}{1.04}
\begin{tabularx}{\textwidth}{@{}p{0.19\textwidth}Xrrrrr@{}}
\toprule
\multicolumn{7}{@{}l}{\textit{Panel A: Source and condition coverage}} \\
\midrule
 & \multicolumn{2}{c}{\textbf{News}} & \multicolumn{2}{c}{\textbf{Reddit}} & \multicolumn{2}{c}{\textbf{Total}} \\
\cmidrule(lr){2-3}\cmidrule(lr){4-5}\cmidrule(lr){6-7}
\textbf{Condition} & \textbf{N} & \textbf{Stigma+} & \textbf{N} & \textbf{Stigma+} & \textbf{N} & \textbf{Stigma+} \\
\midrule
Overall & 249 & 38 & 221 & 49 & 470 & 87 \\
ADHD & 42 & 6 & 35 & 7 & 77 & 13 \\
Alcoholism & 37 & 4 & 41 & 8 & 78 & 12 \\
Autism & 36 & 1 & 42 & 15 & 78 & 16 \\
BPD & 47 & 10 & 33 & 10 & 80 & 20 \\
Depression & 38 & 0 & 41 & 4 & 79 & 4 \\
Schizophrenia & 49 & 17 & 29 & 5 & 78 & 22 \\
\midrule
\multicolumn{7}{@{}l}{\textit{Panel B: Source, condition, and stigma mode}} \\
\midrule
\textbf{Level} & \textbf{Group} & \textbf{N} & \textbf{Stigma+} &
\textbf{\shortstack{Expressed/\\enacted}} &
\textbf{\shortstack{Perceived/\\experienced}} &
\textbf{Both} \\
\midrule
Overall & Overall & 470 & 87 & 49 & 15 & 23 \\
\midrule
Source & News & 249 & 38 & 29 & 5 & 4 \\
Source & Reddit & 221 & 49 & 20 & 10 & 19 \\
\midrule
Condition & ADHD & 77 & 13 & 5 & 5 & 3 \\
Condition & Alcoholism & 78 & 12 & 7 & 1 & 4 \\
Condition & Autism & 78 & 16 & 9 & 5 & 2 \\
Condition & BPD & 80 & 20 & 11 & 1 & 8 \\
Condition & Depression & 79 & 4 & 0 & 1 & 3 \\
Condition & Schizophrenia & 78 & 22 & 17 & 2 & 3 \\
\midrule
\multicolumn{7}{@{}l}{\textit{Panel C: Higher-level taxonomy coverage among stigma-present texts}} \\
\midrule
\textbf{Level} & \textbf{Category} & \multicolumn{5}{r}{\textbf{N labels}} \\
\midrule
Domain & Public & \multicolumn{5}{r}{62} \\
Domain & Self & \multicolumn{5}{r}{24} \\
Domain & Structural & \multicolumn{5}{r}{6} \\
Domain & Associative & \multicolumn{5}{r}{1} \\
\midrule
Public-stigma subfactor & Cognitive & \multicolumn{5}{r}{54} \\
Public-stigma subfactor & Affective & \multicolumn{5}{r}{9} \\
Public-stigma subfactor & Behavioral & \multicolumn{5}{r}{5} \\
\midrule
\multicolumn{7}{@{}l}{\textit{Panel D: Fine-grained public-stigma components among stigma-present texts}} \\
\midrule
\textbf{Subfactor} & \textbf{Component} & \multicolumn{5}{r}{\textbf{N labels}} \\
\midrule
Cognitive & Moral or character flaw & \multicolumn{5}{r}{14} \\
Cognitive & Incompetence & \multicolumn{5}{r}{14} \\
Cognitive & Uncontrollable & \multicolumn{5}{r}{11} \\
Cognitive & Trivialization/minimization & \multicolumn{5}{r}{9} \\
Cognitive & Pity/condescension & \multicolumn{5}{r}{8} \\
Cognitive & Dangerousness & \multicolumn{5}{r}{7} \\
Cognitive & Immutability & \multicolumn{5}{r}{5} \\
Cognitive & Segregation endorsement & \multicolumn{5}{r}{4} \\
Cognitive & Controllable & \multicolumn{5}{r}{3} \\
Cognitive & Personal responsibility/blame & \multicolumn{5}{r}{2} \\
Cognitive & Contamination & \multicolumn{5}{r}{1} \\
\midrule
Affective & Anger/irritability/frustration & \multicolumn{5}{r}{4} \\
Affective & Disgust & \multicolumn{5}{r}{4} \\
Affective & Fear & \multicolumn{5}{r}{2} \\
\midrule
Behavioral & Bullying/mocking & \multicolumn{5}{r}{4} \\
Behavioral & Social distancing & \multicolumn{5}{r}{1} \\
\bottomrule
\end{tabularx}

\caption{Expanded composition of the final benchmark. Panel A reports source-condition coverage using raw text counts. Panel B reports source, condition, and stigma-mode coverage. Panels C and D report taxonomy labels among the 87 stigma-present texts. Counts in Panels C and D refer to label assignments rather than mutually exclusive texts, because stigma-present texts could receive multiple labels. These statistics describe benchmark composition rather than real-world stigma prevalence.}
\label{tab:benchmark_composition_expanded}
\end{table*}

\clearpage

\section{Qualitative model error examples}
\label{app:error_examples_full_model}

Qualitative examples were selected from a systematic error-audit file created using predefined decision rules. The audit file grouped errors by text ID, identified persistent false positives and false negatives across model--prompt combinations, and recorded within-model prompt transitions where added prompt information corrected or introduced an error. The final examples were then manually inspected to assign qualitative error patterns.


\begin{table*}[h]
\centering
\small
\renewcommand{\arraystretch}{1.12}
\begin{tabularx}{\textwidth}{@{}p{0.18\textwidth}p{0.25\textwidth}X@{}}
\toprule
\textbf{Source} & \textbf{Cue triggering overprediction} & \textbf{Interpretation} \\
\midrule

Reddit / schizophrenia
& Risk or causal association
& The text asks about a possible causal or risk association between cannabis use and schizophrenia. It mentions a negative outcome, but does not direct a stigmatizing attribution, treatment, or expression toward people with schizophrenia. \\
\multicolumn{3}{@{}p{\textwidth}@{}}{\textbf{Paraphrased text.} \textit{``someone told me thc only takes 30 days to leave your system no matter what? also heard that if you start smoking weed young you'll definitely get schizophrenia \textbf{schizophrenia}''}} \\

\midrule

Reddit / borderline personality disorder
& Clinical and interpersonal-risk framing
& This example illustrates a difficult threshold case. The passage contains negative, clinically loaded descriptions of BPD symptoms, but the final adjudicated label was stigma-absent. Models treated the diagnostic and interpersonal-risk framing as sufficient evidence of stigma. \\
\multicolumn{3}{@{}p{\textwidth}@{}}{\textbf{Paraphrased text.} \textit{``people with \textbf{borderline personality disorder }(\textbf{bpd}) deal with really intense emotional swings that can put them in a really desperate headspace. its kind of like when you feel cornered - you lash out at whoever is closest to you that makes you feel threatened. and the thing is, the people who know you best are actually the most vulnerable because sufferers of \textbf{bpd} can use what they know about you against you. so the ones who care the most tend to get hurt the most.''}} \\

\midrule

News / depression
& Mental health crisis context
& The passage links student behavior to a broader mental health crisis, but it frames disruptive behavior as unmet need rather than as a stigmatizing attribution toward people with depression. Models may have over-relied on negative behavioral language such as ``outbursts'', ``lashing out'', and ``disruptive''. \\
\multicolumn{3}{@{}p{\textwidth}@{}}{\textbf{Full text.} \textit{``The spike in outbursts, fidgeting and nonstop chatter in classrooms coincides with a national mental health crisis. American pediatricians during the pandemic went as far as to call it a national emergency. A parents with kids under 18 said they are ``extremely or very worried that their children might struggle with anxiety or \textbf{depression} at some point.'' Teachers are under pressure to help students make up for what they haven't learned, and kids feel their stress. The combination has led to a pronounced rise in students expressing anger or sadness by lashing out and being disruptive at school. ``Anytime we've got a chaotic structure and high expectations of ourselves, it's very easy for that to come out in disruptions,'' said Kent McIntosh, an associate professor in the College of Education at the University of Oregon. ``What on the outside can look like defiant behavior is sometimes a student's best bet to get needs met in the moment.''}} \\

\bottomrule
\end{tabularx}
\caption{Selected false-positive examples. These are stigma-absent texts that one or more models predicted as stigma-present. Reddit examples have been paraphrased to preserve user privacy; news examples are reproduced in full.}
\label{tab:false_positive_examples}
\end{table*}

\begin{table*}[h]
\centering
\small
\renewcommand{\arraystretch}{1.12}
\begin{tabularx}{\textwidth}{@{}p{0.18\textwidth}p{0.25\textwidth}X@{}}
\toprule
\textbf{ID / source} & \textbf{Cue missed by models} & \textbf{Interpretation} \\
\midrule

Reddit / alcoholism
& Self-directed shame and guilt
& The broader tone is hopeful and recovery-oriented, but the passage contains self-directed shame and guilt related to alcoholism. Models often missed this case, suggesting that self-stigma may be under-detected when embedded in supportive, religious, or non-toxic discourse. \\
\multicolumn{3}{@{}p{\textwidth}@{}}{\textbf{Full text.} \textit{``heading to church for the first time in a while and i'm actually really looking forward to it. i'd been staying away because i felt so ashamed about my \textbf{alcoholism} but i've realised that's kind of the whole reason i should go. woke up this morning and just prayed for strength and a fresh start. feeling hopeful. hope everyone has a good day''}} \\

\midrule

Reddit / autism
& Structural exclusion / being left behind
& The passage expresses concern about structural exclusion and unequal access to support services for autistic adults and people with disabilities. Models may have missed this case because it appears in advocacy-oriented and institutional language rather than in overtly hostile or derogatory language. \\
\multicolumn{3}{@{}p{\textwidth}@{}}{\textbf{Full text.} \textit{``Two \textbf{autistic} adults here. We rely on in house daily support. The pandemic really showed how essential these kinds of community-based services are for people with disabilities. There's a real opportunity right now with the COVID-19 relief package to actually fund these services properly but honestly we're scared disabled people are going to get overlooked again like always. The Autistic Self Advocacy Network (disability rights advocacy group) has been tracking COVID-19 deaths in care institutions since early 2020, and this research just confirms what we already knew - government decisions literally determine whether people like us get the support we need to survive and live our lives.''}} \\

\midrule

Reddit / ADHD
& Invalidation of diagnosis / ADHD treated as an excuse
& The passage was adjudicated as stigma-present because it describes a family context in which ADHD is dismissed or invalidated: the poster reports that their stepfather has never acknowledged their ADHD and that their mother says it is ``an excuse.'' Models may have missed this case because the stigma is embedded in a first-person help-seeking post rather than expressed as overt abuse or toxic language. \\
\multicolumn{3}{@{}p{\textwidth}@{}}{\textbf{Full text.} \textit{``tl;dr: strict home environment is causing me to have regular meltdowns. basically the pressure at home is so intense that i end up breaking down in the bathroom every couple of weeks. the expectations are really high - grades have to be near perfect, chores can't slip even once, screen time is capped and has to be timed, and i can only use my laptop in a shared space with the screen visible to everyone. i can't go out without giving my parents my exact route in advance, and i'm still not allowed to get my learner's permit so i'm getting around on a bike. there's a lot of other rules on top of all that. the hardest part is that my stepdad completely dismisses my \textbf{adhd} diagnosis that i've had since i was a kid, and my mom says i just use it as an excuse. just needed to get that out - does anyone have any advice?''}} \\

\bottomrule
\end{tabularx}
\caption{Selected false-negative examples. These are stigma-present texts that one or more models predicted as stigma-absent. Reddit examples have been paraphrased to preserve user privacy.}
\label{tab:false_negative_examples}
\end{table*}

\twocolumn

\newpage

\section{Human annotation agreement} 



\label{app:human_agreement}
\begin{figure}[t]
\centering
\includegraphics[width=.82\columnwidth]{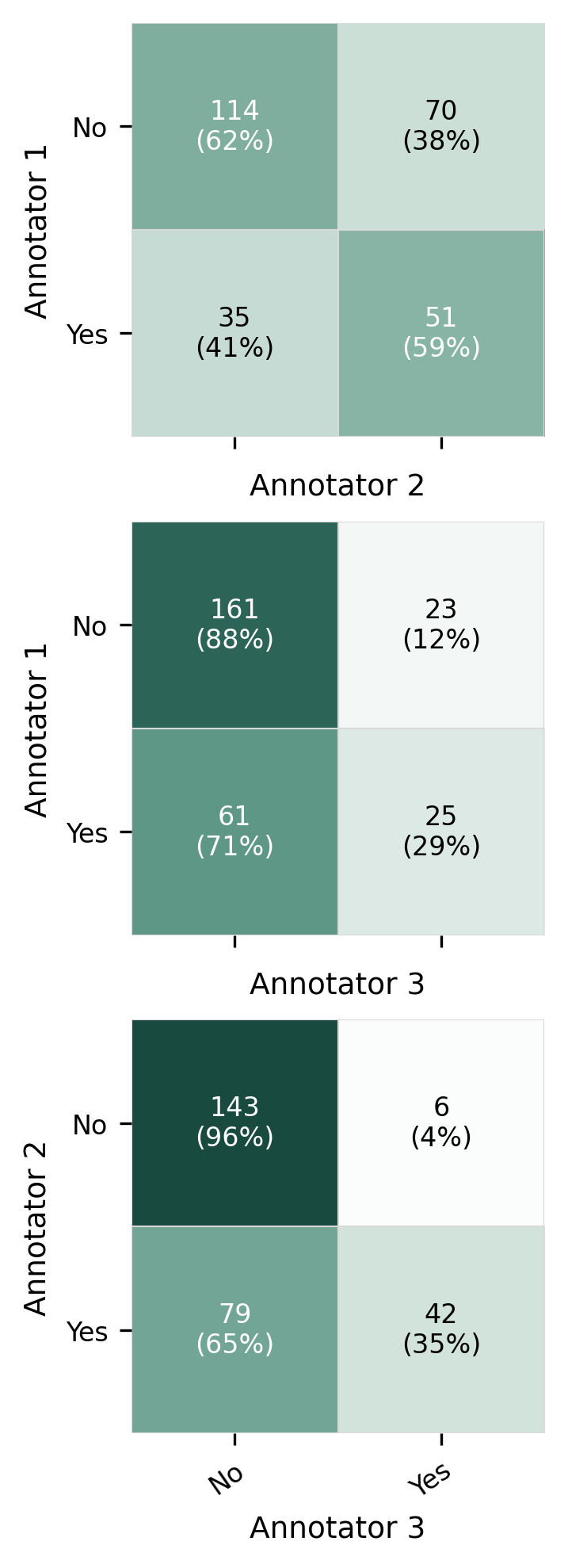}
\caption{Pairwise agreement matrices for stigma presence.}
\label{fig:cm_stigma_present}
\end{figure}

\begin{figure}[t]
\centering
\includegraphics[width=\columnwidth]{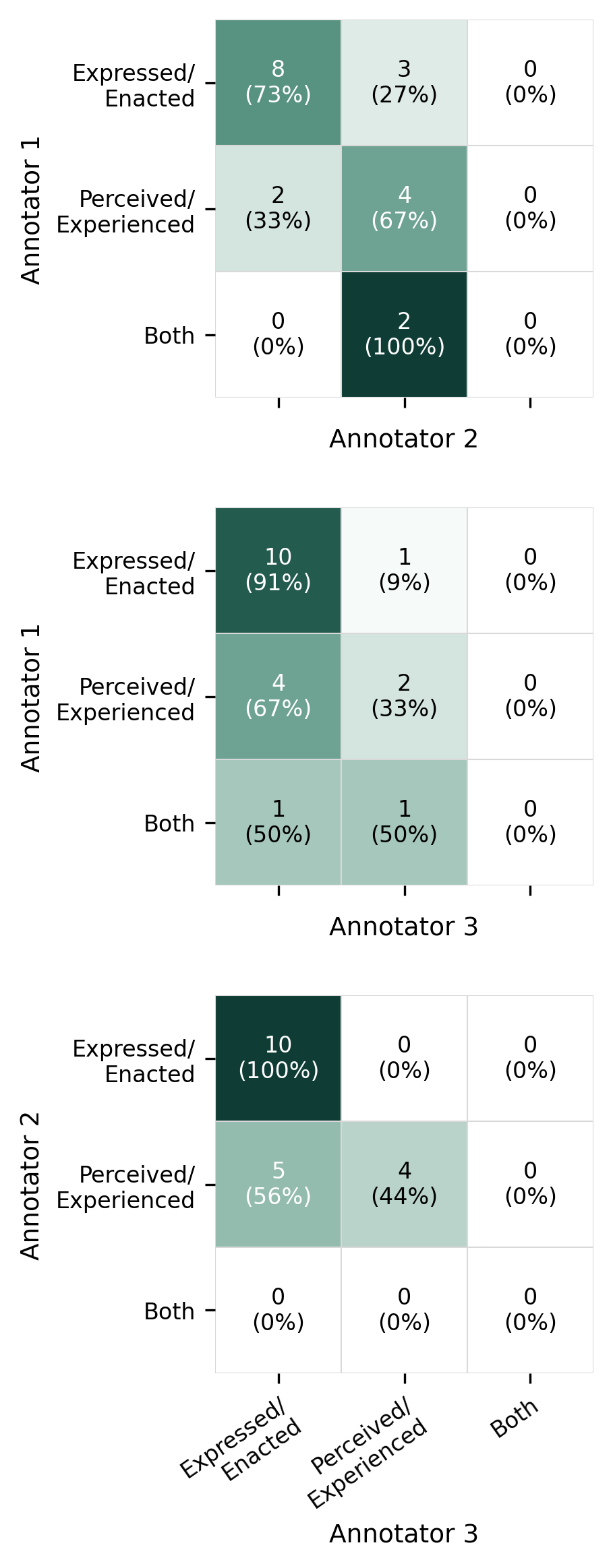}
\caption{Pairwise agreement matrices for stigma mode among the items that all annotators judged as stigma-present.}
\label{fig:cm_mode}
\end{figure}

We first report agreement for all annotation rounds combined. Binary stigma-presence agreement across all items was moderate (exact agreement = .66; AC1 = .40). Agreement on the stigma mode category was also modest when no-stigma was included as a category (exact agreement = .61; AC1 = .54), and among the 19 items that all annotators judged as stigma-present (exact agreement = .51; AC1 = .67). Figures~\ref{fig:cm_stigma_present} and~\ref{fig:cm_mode} show agreement matrices for each annotator pair. 

There was also variation in agreement across rounds (Table \ref{tab:iaa_by_round}). In Round 2, the three experienced annotators achieved an exact agreement of .62 (AC1 = .62) on stigma presence. In Round 3, which introduced a new annotator, agreement fell to .39 (AC1 = .22). Pairwise comparisons suggest this decline was driven by the new annotator, whose agreement with the two returning annotators was low (AC1 = .08 and .11). The new annotator also labeled 66.7\% of items as stigma-present, compared with 30.0\% and 24.7\% for the returning annotators.

When pooled across rounds, pairwise results also suggested threshold differences across annotators. A1–A2 agreement on stigma presence was lower (exact agreement = .61; AC1 = .26) than A1–A3 (exact agreement = .69; AC1 = .50) or A2–A3 (exact agreement = .69; AC1 = .45). Disagreements were resolved through multiple rounds of discussion and adjudication, which led to revisions of the annotation guidelines.

Because subtype and domain labels are only meaningful for stigma-present items, these analyses were restricted to the items that all annotators marked as stigma-present. Agreement was relatively high for domain labels (macro pairwise AC1 = .82) and public-stigma subfactors (macro pairwise AC1 = .89). These estimates should be interpreted cautiously, however, because the subset was small and some labels were rare, making it easier for annotators to agree that a label was absent.

\begin{table}[ht]
\centering
\small
\setlength{\tabcolsep}{4pt}
\begin{tabular}{llrrr}
\toprule
\textbf{Round} & \textbf{Pair} & \textbf{\textit{n}} & \textbf{Exact} & \textbf{AC1} \\
\midrule
R2 & A1--A2 & 120 & .683 & .521 \\
   & A1--A3 & 120 & .700 & .514 \\
   & A2--A3 & 120 & .850 & .805 \\
   & \textit{Overall} & 120 & .617 & .622 \\
\midrule
R3 & A1--A2 & 150 & .693 & .491 \\
   & A1--A3 & 150 & .540 & .081 \\
   & A2--A3 & 150 & .553 & .113 \\
   & \textit{Overall} & 150 & .393 & .220 \\
\bottomrule
\end{tabular}
\caption{Pairwise and overall agreement on stigma presence by annotation round. Exact = exact agreement; AC1 = Gwet's AC1. The A3 position was held by a psychology professor in Round 2 and by a psychology PhD student in Round 3.}
\label{tab:iaa_by_round}
\end{table}





\onecolumn

\section{Detailed scores for model predictions} 
\label{app:detailed_scores}

\begin{table*}[th!]
\centering
\footnotesize
\setlength{\tabcolsep}{3pt}
\begin{tabular}{llcccccccc}
\toprule
& & & \multicolumn{3}{c}{\textbf{Binary}} & \multicolumn{3}{c}{\textbf{Macro}} \\
\cmidrule(lr){4-6} \cmidrule(lr){7-9}
\textbf{Strategy} & \textbf{Model} & \textbf{Acc} & \textbf{P} & \textbf{R} & \textbf{F1} & \textbf{P} & \textbf{R} & \textbf{F1} \\
\midrule
 
\multirow{4}{*}{\texttt{baseline}}
 & Claude Sonnet 4.6 & .67 \scriptsize[.62, .71] & .31 \scriptsize[.24, .37] & .66 \scriptsize[.55, .75] & .42 \scriptsize[.35, .49] & .60 \scriptsize[.57, .64] & .66 \scriptsize[.61, .72] & .59 \scriptsize[.55, .64] \\
 & Deepseek v4 Flash & .70 \scriptsize[.66, .75] & .27 \scriptsize[.19, .35] & .34 \scriptsize[.25, .45] & .30 \scriptsize[.22, .39] & .55 \scriptsize[.51, .60] & .57 \scriptsize[.51, .62] & .56 \scriptsize[.51, .61] \\
 & Gemini 3 Flash    & .73 \scriptsize[.69, .78] & .39 \scriptsize[.31, .46] & .74 \scriptsize[.64, .82] & \textbf{.51} \scriptsize[.43, .58] & \textbf{\textit{.65}} \scriptsize[.61, .69] & .73 \scriptsize[.68, .79] & \textbf{.66} \scriptsize[.61, .71] \\
 & GPT-5.5           & .57 \scriptsize[.52, .61] & .28 \scriptsize[.23, .34] & .86 \scriptsize[.78, .93] & .42 \scriptsize[.36, .49] & .61 \scriptsize[.58, .64] & .68 \scriptsize[.63, .72] & .54 \scriptsize[.49, .58] \\
\midrule
 
\multirow{4}{*}{\texttt{definition}}
 & Claude Sonnet 4.6 & .70 \scriptsize[.66, .74] & .32 \scriptsize[.25, .39] & .54 \scriptsize[.44, .65] & .40 \scriptsize[.32, .48] & .60 \scriptsize[.56, .64] & .64 \scriptsize[.58, .69] & .60 \scriptsize[.55, .65] \\
 & Deepseek v4 Flash & .63 \scriptsize[.58, .67] & .27 \scriptsize[.21, .33] & .60 \scriptsize[.49, .70] & .37 \scriptsize[.30, .44] & .57 \scriptsize[.53, .61] & .61 \scriptsize[.56, .67] & .55 \scriptsize[.50, .60] \\
 & Gemini 3 Flash    & .73 \scriptsize[.70, .77] & .38 \scriptsize[.30, .46] & .68 \scriptsize[.58, .78] & .49 \scriptsize[.40, .56] & .64 \scriptsize[.60, .69] & .71 \scriptsize[.66, .76] & .65 \scriptsize[.61, .70] \\
 & GPT-5.5           & .69 \scriptsize[.65, .74] & .34 \scriptsize[.28, .41] & .72 \scriptsize[.62, .81] & .46 \scriptsize[.39, .54] & .63 \scriptsize[.59, .67] & .70 \scriptsize[.65, .75] & .62 \scriptsize[.58, .67] \\
\midrule
 
\multirow{4}{*}{\texttt{examples}}
 & Claude Sonnet 4.6 & .72 \scriptsize[.68, .76] & .35 \scriptsize[.27, .43] & .56 \scriptsize[.46, .67] & .43 \scriptsize[.35, .51] & .62 \scriptsize[.57, .66] & .66 \scriptsize[.61, .72] & .62 \scriptsize[.57, .67] \\
 & Deepseek v4 Flash & .65 \scriptsize[.61, .69] & .28 \scriptsize[.22, .35] & .57 \scriptsize[.47, .68] & .38 \scriptsize[.30, .45] & .58 \scriptsize[.54, .62] & .62 \scriptsize[.56, .68] & .57 \scriptsize[.52, .61] \\
 & Gemini 3 Flash    & .67 \scriptsize[.62, .71] & .33 \scriptsize[.26, .39] & .75 \scriptsize[.65, .84] & .45 \scriptsize[.38, .52] & .62 \scriptsize[.59, .66] & .70 \scriptsize[.65, .75] & .61 \scriptsize[.56, .65] \\
 & GPT-5.5           & .65 \scriptsize[.60, .69] & .33 \scriptsize[.27, .39] & \textbf{\textit{.89}} \scriptsize[.81, .95] & .48 \scriptsize[.41, .55] & .64 \scriptsize[.61, .68] & \textbf{\textit{.74}} \scriptsize[.69, .78] & .61 \scriptsize[.56, .65] \\
\midrule
 
\multirow{4}{*}{\texttt{examples+def.}}
 & Claude Sonnet 4.6 & .73 \scriptsize[.69, .77] & .35 \scriptsize[.28, .44] & .54 \scriptsize[.44, .64] & .43 \scriptsize[.35, .51] & .62 \scriptsize[.57, .66] & .66 \scriptsize[.60, .71] & .63 \scriptsize[.58, .67] \\
 & Deepseek v4 Flash & .64 \scriptsize[.60, .69] & .27 \scriptsize[.21, .34] & .56 \scriptsize[.46, .66] & .37 \scriptsize[.29, .44] & .57 \scriptsize[.53, .61] & .61 \scriptsize[.56, .67] & .56 \scriptsize[.51, .60] \\
 & Gemini 3 Flash    & .68 \scriptsize[.63, .72] & .33 \scriptsize[.26, .40] & .71 \scriptsize[.62, .81] & .45 \scriptsize[.37, .52] & .62 \scriptsize[.58, .66] & .69 \scriptsize[.64, .74] & .61 \scriptsize[.56, .66] \\
 & GPT-5.5           & .66 \scriptsize[.62, .71] & .33 \scriptsize[.27, .39] & .79 \scriptsize[.70, .88] & .47 \scriptsize[.39, .53] & .63 \scriptsize[.60, .67] & .71 \scriptsize[.66, .76] & .61 \scriptsize[.56, .66] \\
\midrule
 
\multirow{4}{*}{\texttt{ex.+def.+rule}}
 & Claude Sonnet 4.6 & \textbf{\textit{.78}} \scriptsize[.74, .81] & \textbf{\textit{.40}} \scriptsize[.29, .51] & .36 \scriptsize[.26, .45] & .38 \scriptsize[.28, .47] & .63 \scriptsize[.57, .68] & .62 \scriptsize[.56, .67] & .62 \scriptsize[.57, .67] \\
 & Deepseek v4 Flash & .71 \scriptsize[.67, .75] & .31 \scriptsize[.24, .40] & .49 \scriptsize[.39, .60] & .38 \scriptsize[.30, .46] & .59 \scriptsize[.55, .63] & .62 \scriptsize[.57, .68] & .60 \scriptsize[.55, .64] \\
 & Gemini 3 Flash    & .73 \scriptsize[.69, .77] & .36 \scriptsize[.28, .45] & .55 \scriptsize[.45, .65] & .43 \scriptsize[.35, .52] & .62 \scriptsize[.58, .67] & .66 \scriptsize[.61, .72] & .63 \scriptsize[.58, .68] \\
 & GPT-5.5           & .67 \scriptsize[.62, .71] & .33 \scriptsize[.26, .39] & .75 \scriptsize[.65, .84] & .45 \scriptsize[.38, .52] & .62 \scriptsize[.58, .66] & .70 \scriptsize[.64, .75] & .61 \scriptsize[.56, .65] \\
\bottomrule
\end{tabular}
\caption{Stigma presence detection results (accuracy, binary and macro-averaged precision, recall, and F1) across experimental settings and models. Binary metrics are reported in terms of the positive class (``stigma present''). Bracketed values are 95\% confidence intervals obtained from bootstrapping (2000 resamples). Best results across all models and conditions are in \textbf{bold}; \textit{\textbf{bold italics}}  means that the difference with the next-best result for a particular metric is also statistically significant (per-instance McNemar test, $\chi^2$ with continuity correction, p < 0.05).}
\label{tab:stigma_presence_cis}

\end{table*}

\Cref{tab:stigma_presence_cis,tab:mode_detailed,tab:domain_detailed,tab:type_detailed,tab:cognitive_subtype_detailed,tab:affective_behavioral_subtype_detailed} show detailed results for detecting stigma, as well as for mode, domain, stigma type and subtype predictions. \Cref{tab:by_condition_full} further breaks down stigma detection performance of Claude Sonnet 4.6 by mental health condition.

\begin{table*}[h]
\centering
\small
\setlength{\tabcolsep}{4pt}
\begin{tabular}{llcccc}
\toprule
\textbf{Strategy} & \textbf{Model} & \textbf{Accuracy} & \textbf{Precision} & \textbf{Recall} & \textbf{F1} \\
\midrule
 
\multirow{4}{*}{\texttt{baseline}}
 & Claude Sonnet 4.6 & 0.563 \scriptsize[.45, .67] & 0.788 \scriptsize[.73, .85] & 0.868 \scriptsize[.81, .92] & 0.809 \scriptsize[.75, .85] \\
 & Deepseek v4 Flash & 0.575 \scriptsize[.47, .68] & 0.839 \scriptsize[.77, .91] & 0.698 \scriptsize[.62, .77] & 0.759 \scriptsize[.69, .82] \\
 & Gemini 3 Flash & 0.575 \scriptsize[.47, .68] & 0.804 \scriptsize[.74, .87] & 0.826 \scriptsize[.77, .88] & 0.792 \scriptsize[.73, .84] \\
 & GPT-5.5 & 0.621 \scriptsize[.52, .71] & 0.842 \scriptsize[.78, .91] & 0.831 \scriptsize[.76, .90] & 0.837 \scriptsize[.79, .88] \\
\midrule
 
\multirow{4}{*}{\texttt{definition}}
 & Claude Sonnet 4.6 & 0.540 \scriptsize[.43, .64] & 0.785 \scriptsize[.72, .84] & 0.848 \scriptsize[.79, .90] & 0.798 \scriptsize[.75, .84] \\
 & Deepseek v4 Flash & 0.632 \scriptsize[.53, .74] & 0.848 \scriptsize[.78, .91] & 0.809 \scriptsize[.74, .88] & 0.826 \scriptsize[.77, .88] \\
 & Gemini 3 Flash & 0.586 \scriptsize[.48, .69] & 0.833 \scriptsize[.77, .89] & 0.799 \scriptsize[.74, .85] & 0.787 \scriptsize[.72, .84] \\
 & GPT-5.5 & 0.609 \scriptsize[.51, .71] & 0.825 \scriptsize[.76, .89] & 0.850 \scriptsize[.78, .92] & 0.836 \scriptsize[.79, .88] \\
\midrule
 
\multirow{4}{*}{\texttt{examples}}
 & Claude Sonnet 4.6 & 0.552 \scriptsize[.45, .66] & 0.774 \scriptsize[.71, .83] & 0.869 \scriptsize[.81, .92] & 0.804 \scriptsize[.75, .85] \\
 & Deepseek v4 Flash & 0.575 \scriptsize[.47, .68] & 0.862 \scriptsize[.79, .93] & 0.704 \scriptsize[.64, .78] & 0.773 \scriptsize[.71, .83] \\
 & Gemini 3 Flash & 0.575 \scriptsize[.47, .68] & 0.838 \scriptsize[.77, .90] & 0.812 \scriptsize[.76, .87] & 0.806 \scriptsize[.75, .86] \\
 & GPT-5.5 & 0.701 \scriptsize[.60, .79] & 0.910 \scriptsize[.86, .95] & 0.852 \scriptsize[.79, .91] & 0.878 \scriptsize[.83, .92] \\
\midrule

\multirow{4}{*}{\texttt{examples+definition}}
 & Claude Sonnet 4.6 & 0.517 \scriptsize[.41, .62] & 0.809 \scriptsize[.74, .87] & 0.779 \scriptsize[.72, .84] & 0.761 \scriptsize[.70, .82] \\
 & Deepseek v4 Flash & 0.598 \scriptsize[.49, .69] & 0.856 \scriptsize[.79, .92] & 0.761 \scriptsize[.70, .83] & 0.801 \scriptsize[.75, .85] \\
 & Gemini 3 Flash & 0.552 \scriptsize[.45, .66] & 0.842 \scriptsize[.78, .90] & 0.737 \scriptsize[.68, .79] & 0.741 \scriptsize[.67, .81] \\
 & GPT-5.5 & 0.690 \scriptsize[.59, .78] & \textbf{0.910} \scriptsize[.85, .96] & 0.836 \scriptsize[.77, .90] & 0.871 \scriptsize[.82, .92] \\
\midrule
 
\multirow{4}{*}{\texttt{+rules}}
 & Claude Sonnet 4.6 & 0.678 \scriptsize[.57, .77] & 0.806 \scriptsize[.74, .87] & 0.835 \scriptsize[.77, .89] & 0.801 \scriptsize[.73, .86] \\
 & Deepseek v4 Flash & 0.575 \scriptsize[.47, .68] & 0.862 \scriptsize[.79, .92] & 0.755 \scriptsize[.68, .83] & 0.804 \scriptsize[.74, .85] \\
 & Gemini 3 Flash & 0.805 \scriptsize[.72, .87] & 0.851 \scriptsize[.79, .91] & \textbf{0.952} \scriptsize[.91, .98] & 0.896 \scriptsize[.84, .94] \\
 & GPT-5.5 & \textbf{0.816} \scriptsize[.74, .90] & 0.903 \scriptsize[.86, .95] & 0.940 \scriptsize[.88, .99] & \textbf{0.921} \scriptsize[.88, .96] \\
\bottomrule
\end{tabular}
\caption{Macro-averaged prediction results for stigma mode (n=87). Bracketed values are 95\% bootstrap confidence intervals (1000 resamples).}
\label{tab:mode_detailed}
\end{table*}

\begin{table*}[ht]
\centering
\small
\setlength{\tabcolsep}{4pt}
\begin{tabular}{llcccc}
\toprule
\textbf{Strategy} & \textbf{Model} & \textbf{Accuracy} & \textbf{Precision} & \textbf{Recall} & \textbf{F1} \\
\midrule
 
\multirow{4}{*}{\texttt{baseline}}
 & Claude Sonnet 4.6 & 0.529 \scriptsize[.43, .63] & 0.569 \scriptsize[.40, .73] & \textbf{0.988} \scriptsize[.73, 1.00] & 0.706 \scriptsize[.49, .83] \\
 & Deepseek v4 Flash & 0.483 \scriptsize[.38, .59] & 0.461 \scriptsize[.39, .53] & 0.690 \scriptsize[.65, .73] & 0.532 \scriptsize[.46, .60] \\
 & Gemini 3 Flash & 0.517 \scriptsize[.41, .62] & 0.462 \scriptsize[.40, .53] & 0.960 \scriptsize[.69, .98] & 0.559 \scriptsize[.47, .64] \\
 & GPT-5.5 & 0.713 \scriptsize[.62, .80] & 0.787 \scriptsize[.49, .85] & 0.982 \scriptsize[.71, 1.00] & 0.854 \scriptsize[.54, .90] \\
\midrule
 
\multirow{4}{*}{\texttt{definition}}
 & Claude Sonnet 4.6 & 0.563 \scriptsize[.45, .67] & 0.527 \scriptsize[.40, .72] & \textbf{0.988} \scriptsize[.73, 1.00] & 0.662 \scriptsize[.49, .82] \\
 & Deepseek v4 Flash & 0.575 \scriptsize[.47, .68] & 0.481 \scriptsize[.41, .55] & 0.723 \scriptsize[.70, .75] & 0.572 \scriptsize[.49, .63] \\
 & Gemini 3 Flash & 0.598 \scriptsize[.49, .70] & 0.512 \scriptsize[.45, .58] & 0.976 \scriptsize[.71, .99] & 0.601 \scriptsize[.51, .68] \\
 & GPT-5.5 & 0.724 \scriptsize[.63, .82] & 0.678 \scriptsize[.50, .84] & 0.967 \scriptsize[.69, .99] & 0.780 \scriptsize[.55, .89] \\
\midrule
 
\multirow{4}{*}{\texttt{examples}}
 & Claude Sonnet 4.6 & 0.552 \scriptsize[.45, .66] & 0.574 \scriptsize[.40, .74] & 0.934 \scriptsize[.62, .99] & 0.696 \scriptsize[.48, .82] \\
 & Deepseek v4 Flash & 0.632 \scriptsize[.54, .74] & 0.569 \scriptsize[.46, .72] & 0.879 \scriptsize[.56, .94] & 0.656 \scriptsize[.50, .79] \\
 & Gemini 3 Flash & 0.621 \scriptsize[.52, .71] & 0.504 \scriptsize[.43, .59] & 0.964 \scriptsize[.70, .98] & 0.595 \scriptsize[.49, .69] \\
 & GPT-5.5 & \textbf{0.759} \scriptsize[.67, .84] & \textbf{0.818} \scriptsize[.51, .89] & 0.971 \scriptsize[.70, 1.00] & \textbf{0.877} \scriptsize[.56, .93] \\
\midrule
 
\multirow{4}{*}{\texttt{examples+definition}}
 & Claude Sonnet 4.6 & 0.575 \scriptsize[.46, .68] & 0.494 \scriptsize[.41, .60] & 0.960 \scriptsize[.69, .98] & 0.598 \scriptsize[.47, .71] \\
 & Deepseek v4 Flash & 0.632 \scriptsize[.53, .72] & 0.564 \scriptsize[.48, .66] & 0.923 \scriptsize[.65, .95] & 0.642 \scriptsize[.52, .75] \\
 & Gemini 3 Flash & 0.598 \scriptsize[.49, .70] & 0.504 \scriptsize[.43, .58] & 0.947 \scriptsize[.67, .97] & 0.581 \scriptsize[.49, .67] \\
 & GPT-5.5 & 0.747 \scriptsize[.66, .83] & 0.809 \scriptsize[.50, .87] & 0.945 \scriptsize[.67, .98] & 0.856 \scriptsize[.54, .91] \\
\bottomrule
\end{tabular}
\caption{Macro-averaged results for stigma domain (n=87). Bracketed values are 95\% bootstrap confidence intervals (1000 resamples).}
\label{tab:domain_detailed}

\end{table*}

\begin{table*}[ht]
\centering
\small
\setlength{\tabcolsep}{4pt}
\begin{tabular}{llcccc}
\toprule
\textbf{Strategy} & \textbf{Model} & \textbf{Accuracy} & \textbf{Precision} & \textbf{Recall} & \textbf{F1} \\
\midrule
 
\multirow{4}{*}{\texttt{baseline}}
 & Claude Sonnet 4.6 & 0.386 \scriptsize[.26, .51] & 0.441 \scriptsize[.37, .52] & \textbf{0.795} \scriptsize[.61, .95] & 0.528 \scriptsize[.43, .61] \\
 & Deepseek v4 Flash & 0.351 \scriptsize[.25, .47] & 0.381 \scriptsize[.32, .46] & 0.655 \scriptsize[.44, .88] & 0.441 \scriptsize[.35, .54] \\
 & Gemini 3 Flash & 0.632 \scriptsize[.51, .75] & 0.364 \scriptsize[.30, .44] & 0.593 \scriptsize[.43, .67] & 0.426 \scriptsize[.33, .51] \\
 & GPT-5.5 & 0.596 \scriptsize[.47, .74] & 0.521 \scriptsize[.39, .68] & 0.733 \scriptsize[.52, .91] & \textbf{0.562} \scriptsize[.41, .69] \\
\midrule
 
\multirow{4}{*}{\texttt{definitions}}
 & Claude Sonnet 4.6 & 0.474 \scriptsize[.35, .61] & 0.412 \scriptsize[.32, .51] & 0.670 \scriptsize[.49, .81] & 0.473 \scriptsize[.36, .57] \\
 & Deepseek v4 Flash & 0.456 \scriptsize[.33, .60] & 0.430 \scriptsize[.33, .55] & 0.666 \scriptsize[.41, .87] & 0.489 \scriptsize[.35, .63] \\
 & Gemini 3 Flash & 0.667 \scriptsize[.54, .79] & 0.539 \scriptsize[.33, .76] & 0.621 \scriptsize[.45, .76] & 0.494 \scriptsize[.36, .63] \\
 & GPT-5.5 & 0.561 \scriptsize[.44, .70] & 0.691 \scriptsize[.34, .75] & 0.649 \scriptsize[.46, .80] & 0.528 \scriptsize[.37, .66] \\
\midrule
 
\multirow{4}{*}{\texttt{examples}}
 & Claude Sonnet 4.6 & 0.509 \scriptsize[.39, .63] & 0.458 \scriptsize[.37, .56] & 0.753 \scriptsize[.57, .93] & 0.535 \scriptsize[.41, .63] \\
 & Deepseek v4 Flash & 0.491 \scriptsize[.35, .61] & 0.410 \scriptsize[.34, .49] & 0.655 \scriptsize[.46, .86] & 0.479 \scriptsize[.37, .56] \\
 & Gemini 3 Flash & \textbf{0.737} \scriptsize[.61, .84] & 0.578 \scriptsize[.36, .81] & 0.621 \scriptsize[.45, .76] & 0.530 \scriptsize[.38, .66] \\
 & GPT-5.5 & 0.596 \scriptsize[.46, .74] & 0.501 \scriptsize[.35, .68] & 0.629 \scriptsize[.45, .78] & 0.501 \scriptsize[.37, .62] \\
\midrule

\multirow{4}{*}{\texttt{examples+definitions}}
 & Claude Sonnet 4.6 & 0.526 \scriptsize[.40, .67] & 0.453 \scriptsize[.36, .56] & 0.711 \scriptsize[.53, .88] & 0.523 \scriptsize[.40, .63] \\
 & Deepseek v4 Flash & 0.456 \scriptsize[.33, .60] & 0.420 \scriptsize[.35, .50] & 0.763 \scriptsize[.54, .94] & 0.503 \scriptsize[.40, .61] \\
 & Gemini 3 Flash & 0.719 \scriptsize[.60, .82] & \textbf{0.722} \scriptsize[.32, .81] & 0.548 \scriptsize[.35, .72] & 0.504 \scriptsize[.34, .64] \\
 & GPT-5.5 & 0.614 \scriptsize[.49, .74] & 0.545 \scriptsize[.35, .75] & 0.636 \scriptsize[.43, .79] & 0.522 \scriptsize[.37, .65] \\
\bottomrule
\end{tabular}
\caption{Macro-averaged results for stigma type (Public domain only, n=57). Bracketed values are 95\% bootstrap confidence intervals (1000 resamples).}
\label{tab:type_detailed}
\end{table*}

\begin{table*}[ht]
\centering
\small
\setlength{\tabcolsep}{4pt}
\begin{tabular}{llcccc}
\toprule
\textbf{Strategy} & \textbf{Model} & \textbf{Acc} & \textbf{Pr} & \textbf{Re} & \textbf{F1} \\
\midrule
 
\multirow{4}{*}{\texttt{baseline}}
 & Claude Sonnet 4.6 & 0.14 \scriptsize[.06,.24] & 0.57 \scriptsize[.33,.68] & 0.63 \scriptsize[.39,.76] & 0.49 \scriptsize[.30,.61] \\
 & Deepseek v4 Flash & 0.12 \scriptsize[.04,.22] & 0.36 \scriptsize[.22,.45] & 0.46 \scriptsize[.34,.52] & 0.36 \scriptsize[.24,.41] \\
 & Gemini 3 Flash & 0.18 \scriptsize[.08,.29] & 0.55 \scriptsize[.37,.59] & 0.59 \scriptsize[.40,.68] & 0.48 \scriptsize[.35,.56] \\
 & GPT-5.5 & 0.24 \scriptsize[.14,.37] & 0.58 \scriptsize[.37,.68] & 0.67 \scriptsize[.41,.75] & 0.50 \scriptsize[.36,.61] \\
\midrule
 
\multirow{4}{*}{\texttt{definition}}
 & Claude Sonnet 4.6 & 0.22 \scriptsize[.12,.35] & 0.52 \scriptsize[.34,.61] & 0.59 \scriptsize[.39,.68] & 0.48 \scriptsize[.31,.56] \\
 & Deepseek v4 Flash & 0.18 \scriptsize[.08,.31] & 0.40 \scriptsize[.28,.48] & 0.53 \scriptsize[.35,.61] & 0.38 \scriptsize[.26,.47] \\
 & Gemini 3 Flash & 0.16 \scriptsize[.06,.27] & 0.49 \scriptsize[.31,.54] & 0.51 \scriptsize[.35,.60] & 0.42 \scriptsize[.30,.49] \\
 & GPT-5.5 & 0.22 \scriptsize[.12,.35] & 0.62 \scriptsize[.46,.70] & 0.70 \scriptsize[.43,.79] & 0.56 \scriptsize[.42,.66] \\
\midrule
 
\multirow{4}{*}{\texttt{examples}}
 & Claude Sonnet 4.6 & 0.16 \scriptsize[.06,.27] & 0.47 \scriptsize[.31,.58] & 0.57 \scriptsize[.38,.65] & 0.44 \scriptsize[.30,.52] \\
 & Deepseek v4 Flash & 0.10 \scriptsize[.02,.18] & 0.24 \scriptsize[.17,.34] & 0.45 \scriptsize[.30,.52] & 0.29 \scriptsize[.21,.34] \\
 & Gemini 3 Flash & 0.20 \scriptsize[.10,.33] & 0.53 \scriptsize[.37,.62] & 0.59 \scriptsize[.37,.67] & 0.49 \scriptsize[.34,.57] \\
 & GPT-5.5 & 0.20 \scriptsize[.10,.33] & 0.50 \scriptsize[.34,.63] & 0.52 \scriptsize[.33,.62] & 0.44 \scriptsize[.31,.53] \\
\midrule
 
\multirow{4}{*}{\texttt{examples+definition}}
 & Claude Sonnet 4.6 & 0.24 \scriptsize[.12,.37] & 0.43 \scriptsize[.32,.56] & 0.61 \scriptsize[.40,.69] & 0.46 \scriptsize[.32,.56] \\
 & Deepseek v4 Flash & 0.14 \scriptsize[.04,.24] & 0.36 \scriptsize[.22,.46] & 0.51 \scriptsize[.35,.58] & 0.37 \scriptsize[.25,.44] \\
 & Gemini 3 Flash & 0.18 \scriptsize[.08,.29] & 0.52 \scriptsize[.36,.62] & 0.61 \scriptsize[.40,.69] & 0.49 \scriptsize[.35,.57] \\
 & GPT-5.5 & 0.24 \scriptsize[.14,.37] & 0.58 \scriptsize[.44,.71] & 0.67 \scriptsize[.40,.77] & 0.55 \scriptsize[.40,.66] \\
\bottomrule
\end{tabular}
\caption{Macro-averaged results for the Cognitive stigma subtype (n=49). Bracketed values are 95\% bootstrap confidence intervals (1000 resamples).}
\label{tab:cognitive_subtype_detailed}
\end{table*}

\begin{table*}[ht]
\centering
\footnotesize
\setlength{\tabcolsep}{2pt}
\begin{tabular}{llcccccccc}
\toprule
& & \multicolumn{4}{c}{\textbf{Affective}} & \multicolumn{4}{c}{\textbf{Behavioral}} \\
\cmidrule(lr){3-6} \cmidrule(lr){7-10}
\textbf{Strategy} & \textbf{Model} & \textbf{Ac} & \textbf{Pr} & \textbf{Re} & \textbf{F1} & \textbf{Ac} & \textbf{Pr} & \textbf{Re} & \textbf{F1} \\
\midrule
 
\multirow{4}{*}{\texttt{basel.}}
 & Claude & 0.38 \scriptsize[.12,.75] & 0.67 \scriptsize[.33,.67] & 0.39 \scriptsize[.11,.60] & 0.49 \scriptsize[.17,.63] & 0.80 \scriptsize[.40,1.00] & 1.00 \scriptsize[.50,1.00] & 0.88 \scriptsize[.20,1.00] & 0.93 \scriptsize[.29,1.00] \\
 & Deepseek & 0.12 \scriptsize[.00,.38] & 0.61 \scriptsize[.33,.67] & 0.56 \scriptsize[.19,.89] & 0.55 \scriptsize[.25,.68] & 0.20 \scriptsize[.00,.60] & 0.75 \scriptsize[.00,.83] & 0.62 \scriptsize[.00,.88] & 0.53 \scriptsize[.00,.73] \\
 & Gemini & 0.00 \scriptsize[.00,.00] & 0.00 \scriptsize[.00,.00] & 0.00 \scriptsize[.00,.00] & 0.00 \scriptsize[.00,.00] & 0.80 \scriptsize[.40,1.00] & 1.00 \scriptsize[.50,1.00] & 0.88 \scriptsize[.20,1.00] & 0.93 \scriptsize[.29,1.00] \\
 & GPT & 0.38 \scriptsize[.12,.75] & 0.67 \scriptsize[.33,.67] & 0.56 \scriptsize[.17,.67] & 0.60 \scriptsize[.22,.67] & 0.80 \scriptsize[.40,1.00] & 1.00 \scriptsize[.50,1.00] & 0.88 \scriptsize[.20,1.00] & 0.93 \scriptsize[.29,1.00] \\
\midrule
 
\multirow{4}{*}{\texttt{def.}}
 & Claude & 0.12 \scriptsize[.00,.38] & 0.50 \scriptsize[.00,.56] & 0.19 \scriptsize[.00,.50] & 0.27 \scriptsize[.00,.47] & 0.80 \scriptsize[.40,1.00] & 1.00 \scriptsize[.50,1.00] & 0.88 \scriptsize[.20,1.00] & 0.93 \scriptsize[.29,1.00] \\
 & Deepseek & 0.12 \scriptsize[.00,.38] & 0.78 \scriptsize[.33,.83] & 0.36 \scriptsize[.05,.72] & 0.47 \scriptsize[.08,.70] & 0.60 \scriptsize[.20,1.00] & 1.00 \scriptsize[.50,1.00] & 0.75 \scriptsize[.10,1.00] & 0.83 \scriptsize[.17,1.00] \\
 & Gemini & 0.00 \scriptsize[.00,.00] & 0.33 \scriptsize[.00,.33] & 0.17 \scriptsize[.00,.33] & 0.22 \scriptsize[.00,.33] & 0.80 \scriptsize[.40,1.00] & 1.00 \scriptsize[.50,1.00] & 0.88 \scriptsize[.20,1.00] & 0.93 \scriptsize[.29,1.00] \\
 & GPT & 0.25 \scriptsize[.00,.62] & 0.67 \scriptsize[.00,.67] & 0.28 \scriptsize[.00,.58] & 0.39 \scriptsize[.00,.62] & 0.80 \scriptsize[.40,1.00] & 1.00 \scriptsize[.50,1.00] & 0.88 \scriptsize[.20,1.00] & 0.93 \scriptsize[.29,1.00] \\
\midrule
 
\multirow{4}{*}{\texttt{ex.}}
 & Claude  & 0.25 \scriptsize[.00,.62] & 0.67 \scriptsize[.00,.67] & 0.28 \scriptsize[.00,.52] & 0.39 \scriptsize[.00,.58] & 0.80 \scriptsize[.40,1.00] & 1.00 \scriptsize[.50,1.00] & 0.88 \scriptsize[.20,1.00] & 0.93 \scriptsize[.29,1.00] \\
 & Deepseek & 0.12 \scriptsize[.00,.38] & 0.61 \scriptsize[.33,.67] & 0.56 \scriptsize[.19,.89] & 0.55 \scriptsize[.25,.68] & 0.40 \scriptsize[.00,.80] & 0.50 \scriptsize[.00,.50] & 0.25 \scriptsize[.00,.50] & 0.33 \scriptsize[.00,.50] \\
 & Gemini & 0.12 \scriptsize[.00,.38] & 0.33 \scriptsize[.00,.33] & 0.17 \scriptsize[.00,.33] & 0.22 \scriptsize[.00,.33] & 0.80 \scriptsize[.40,1.00] & 1.00 \scriptsize[.50,1.00] & 0.88 \scriptsize[.20,1.00] & 0.93 \scriptsize[.29,1.00] \\
 & GPT & 0.12 \scriptsize[.00,.38] & 0.33 \scriptsize[.00,.33] & 0.22 \scriptsize[.00,.33] & 0.27 \scriptsize[.00,.33] & 0.80 \scriptsize[.40,1.00] & 1.00 \scriptsize[.50,1.00] & 0.88 \scriptsize[.20,1.00] & 0.93 \scriptsize[.29,1.00] \\
\midrule
 
\multirow{4}{*}{\texttt{ex.def.}}
 & Claude 4.6 & 0.25 \scriptsize[.00,.62] & 0.67 \scriptsize[.00,.67] & 0.28 \scriptsize[.00,.52] & 0.39 \scriptsize[.00,.58] & 0.80 \scriptsize[.40,1.00] & 1.00 \scriptsize[.50,1.00] & 0.88 \scriptsize[.20,1.00] & 0.93 \scriptsize[.29,1.00] \\
 & Deepseek  & 0.12 \scriptsize[.00,.38] & 0.62 \scriptsize[.28,.70] & 0.58 \scriptsize[.33,.83] & 0.49 \scriptsize[.33,.69] & 0.60 \scriptsize[.20,1.00] & 1.00 \scriptsize[.50,1.00] & 0.75 \scriptsize[.10,1.00] & 0.83 \scriptsize[.17,1.00] \\
 & Gemini & 0.12 \scriptsize[.00,.38] & 0.33 \scriptsize[.00,.33] & 0.17 \scriptsize[.00,.33] & 0.22 \scriptsize[.00,.33] & 0.60 \scriptsize[.20,1.00] & 1.00 \scriptsize[.50,1.00] & 0.75 \scriptsize[.10,1.00] & 0.83 \scriptsize[.17,1.00] \\
 & GPT & 0.12 \scriptsize[.00,.38] & 0.33 \scriptsize[.00,.33] & 0.11 \scriptsize[.00,.33] & 0.17 \scriptsize[.00,.33] & 0.80 \scriptsize[.40,1.00] & 1.00 \scriptsize[.50,1.00] & 0.88 \scriptsize[.20,1.00] & 0.93 \scriptsize[.29,1.00] \\
\bottomrule
\end{tabular}
\caption{Macro-averaged results for the Affective (n=8) and Behavioral (n=5) stigma subtypes. Bracketed values are 95\% bootstrap confidence intervals (1000 resamples).}
\label{tab:affective_behavioral_subtype_detailed}
\end{table*}

\begin{table*}[t]
\centering
\footnotesize
\setlength{\tabcolsep}{3pt}
\begin{tabular}{llrrrrrrr}
\toprule
\textbf{Strategy} & \textbf{Condition} & \textbf{N} & \textbf{Stig+} & \textbf{Acc} & \textbf{P} & \textbf{R} & \textbf{F1} & \textbf{Bias} \\
\midrule
\multirow{6}{*}{\texttt{baseline}}
 & ADHD          & 77 & 13 & 0.740 & 0.333 & 0.538 & 0.412 & \cellcolor{red!21} 0.104 \\
 & Alcoholism    & 78 & 12 & 0.654 & 0.241 & 0.583 & 0.341 & \cellcolor{red!44} 0.218 \\
 & Autism        & 78 & 16 & 0.718 & 0.400 & 0.750 & 0.522 & \cellcolor{red!36} 0.179 \\
 & BPD           & 80 & 20 & 0.525 & 0.286 & 0.600 & 0.387 & \cellcolor{red!55} 0.275 \\
 & Depression    & 79 &  4 & 0.848 & 0.250 & 1.000 & 0.400 & \cellcolor{red!30} 0.152 \\
 & Schizophrenia & 78 & 22 & 0.513 & 0.326 & 0.682 & 0.441 & \cellcolor{red!62} 0.308 \\
\midrule
\multirow{6}{*}{\texttt{definition}}
 & ADHD          & 77 & 13 & 0.727 & 0.300 & 0.462 & 0.364 & \cellcolor{red!18} 0.091 \\
 & Alcoholism    & 78 & 12 & 0.731 & 0.286 & 0.500 & 0.364 & \cellcolor{red!23} 0.115 \\
 & Autism        & 78 & 16 & 0.744 & 0.417 & 0.625 & 0.500 & \cellcolor{red!21} 0.103 \\
 & BPD           & 80 & 20 & 0.562 & 0.286 & 0.500 & 0.364 & \cellcolor{red!38} 0.188 \\
 & Depression    & 79 &  4 & 0.873 & 0.250 & 0.750 & 0.375 & \cellcolor{red!20} 0.101 \\
 & Schizophrenia & 78 & 22 & 0.551 & 0.324 & 0.545 & 0.407 & \cellcolor{red!38} 0.192 \\
\midrule
\multirow{6}{*}{\texttt{examples}}
 & ADHD          & 77 & 13 & 0.766 & 0.368 & 0.538 & 0.438 & \cellcolor{red!16} 0.078 \\
 & Alcoholism    & 78 & 12 & 0.756 & 0.231 & 0.250 & 0.240 & \cellcolor{red!3} 0.013 \\
 & Autism        & 78 & 16 & 0.769 & 0.458 & 0.688 & 0.550 & \cellcolor{red!21} 0.103 \\
 & BPD           & 80 & 20 & 0.600 & 0.312 & 0.500 & 0.385 & \cellcolor{red!30} 0.150 \\
 & Depression    & 79 &  4 & 0.924 & 0.333 & 0.500 & 0.400 & \cellcolor{red!5} 0.025 \\
 & Schizophrenia & 78 & 22 & 0.526 & 0.340 & 0.727 & 0.464 & \cellcolor{red!64} 0.321 \\
\midrule
\multirow{6}{*}{\texttt{examples+definition}}
 & ADHD          & 77 & 13 & 0.766 & 0.353 & 0.462 & 0.400 & \cellcolor{red!10} 0.052 \\
 & Alcoholism    & 78 & 12 & 0.821 & 0.417 & 0.417 & 0.417 & 0.000 \\
 & Autism        & 78 & 16 & 0.744 & 0.409 & 0.562 & 0.474 & \cellcolor{red!15} 0.077 \\
 & BPD           & 80 & 20 & 0.588 & 0.290 & 0.450 & 0.353 & \cellcolor{red!28} 0.138 \\
 & Depression    & 79 &  4 & 0.962 & 0.667 & 0.500 & 0.571 & \cellcolor{blue!3} $-$0.013 \\
 & Schizophrenia & 78 & 22 & 0.513 & 0.333 & 0.727 & 0.457 & \cellcolor{red!67} 0.333 \\
\midrule
\multirow{6}{*}{\texttt{examples+definition+rule}}
 & ADHD          & 77 & 13 & 0.792 & 0.364 & 0.308 & 0.333 & \cellcolor{blue!5} $-$0.026 \\
 & Alcoholism    & 78 & 12 & 0.833 & 0.429 & 0.250 & 0.316 & \cellcolor{blue!13} $-$0.064 \\
 & Autism        & 78 & 16 & 0.756 & 0.412 & 0.438 & 0.424 & \cellcolor{red!3} 0.013 \\
 & BPD           & 80 & 20 & 0.650 & 0.300 & 0.300 & 0.300 & 0.000 \\
 & Depression    & 79 &  4 & 0.949 & 0.500 & 0.250 & 0.333 & \cellcolor{blue!5} $-$0.025 \\
 & Schizophrenia & 78 & 22 & 0.705 & 0.476 & 0.455 & 0.465 & \cellcolor{blue!3} $-$0.013 \\
\bottomrule
\end{tabular}
\caption{Condition-level stigma-presence results for Claude Sonnet 4.6 by prompting strategy. N = number of items; Stig+ = gold stigma-positive items. Bias = model's predicted positive rate $-$ gold stigma-positive rate; positive values (red) = overprediction, negative values (blue) = underprediction.}
\label{tab:by_condition_full}
\end{table*}

\twocolumn
\section{Confusion matrices for fine-grained stigma classification}
\label{app:confusion_matrices}

This appendix reports additional diagnostic confusion matrices for fine-grained stigma classification. The matrices are intended as descriptive error analyses rather than stable estimates for rare labels, because several fine-grained categories occurred only a small number of times. Cognitive subtype confusions often involved theoretically related categories, including personal responsibility versus trivialization, controllability versus personal responsibility, segregation endorsement versus incompetence, and dangerousness versus uncontrollability. Affective components were also frequently confused, but the small number of affective examples limits interpretation.

\begin{figure*}[p]
\centering

\includegraphics[width=.88\textwidth]{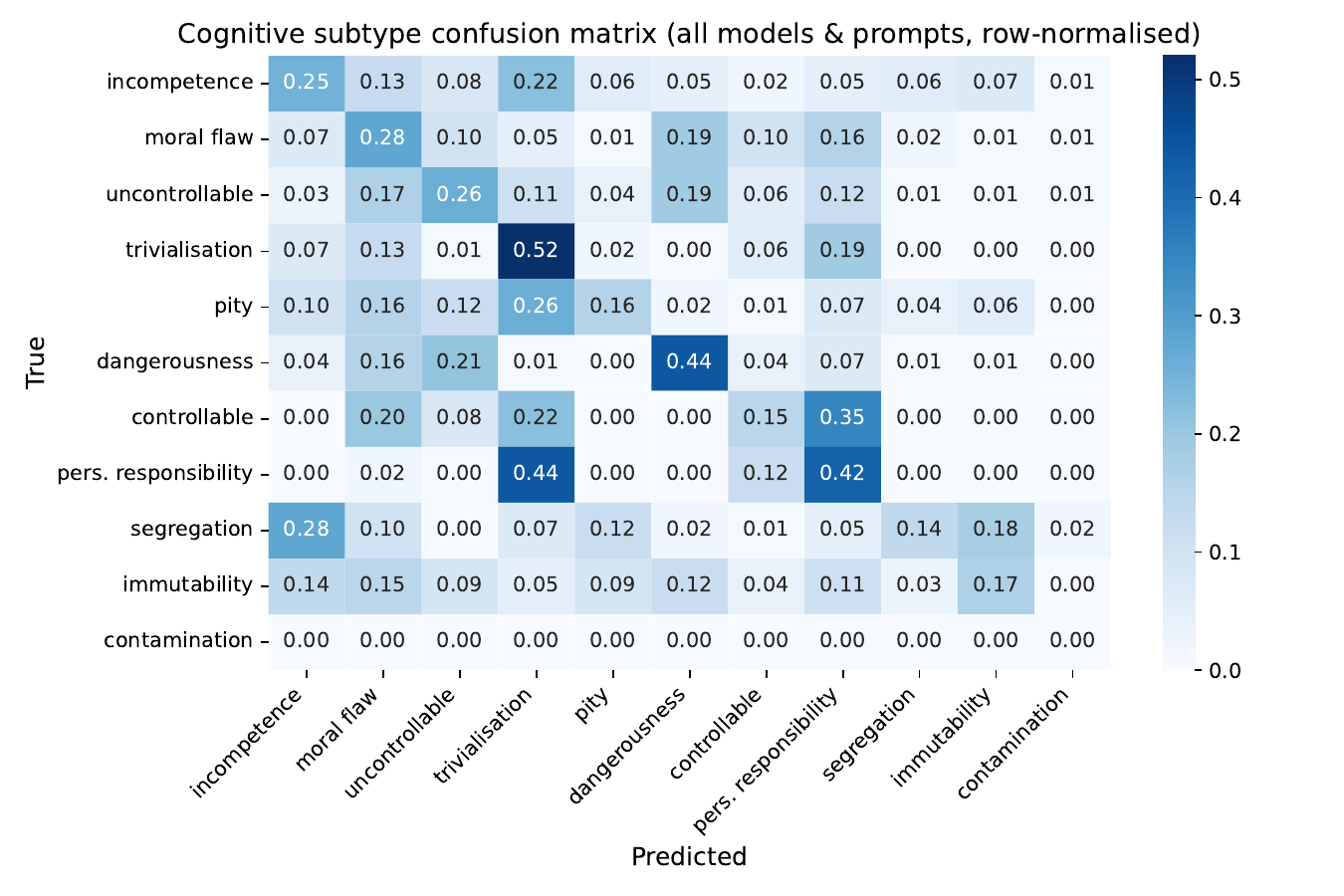}
\caption{Row-normalized model confusion matrix for cognitive public-stigma component prediction.}
\label{fig:model_cm_cognitive}

\vspace{0.5em}

\includegraphics[width=.58\textwidth]{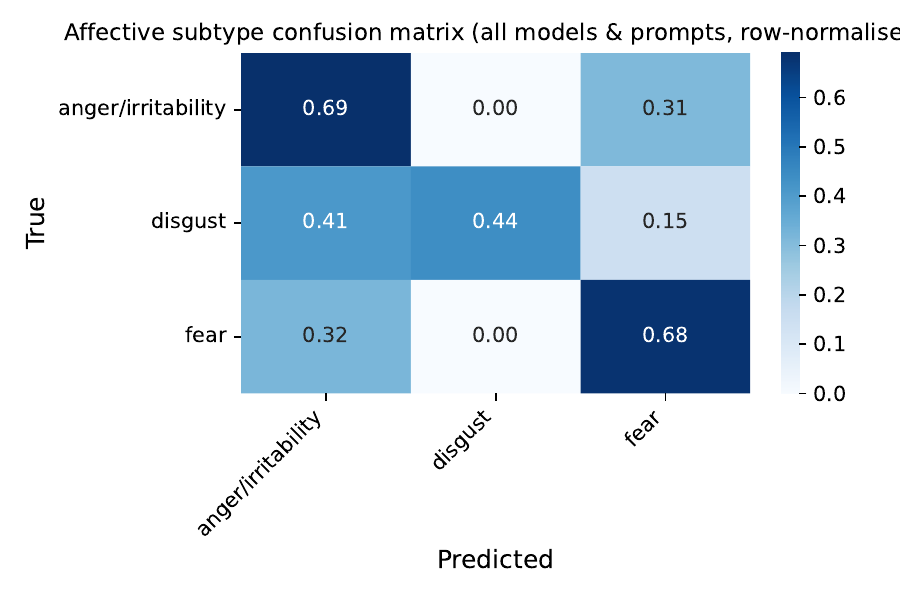}
\caption{Row-normalized model confusion matrix for affective public-stigma component prediction.}
\label{fig:model_cm_affective}

\vspace{0.5em}

\includegraphics[width=.50\textwidth]{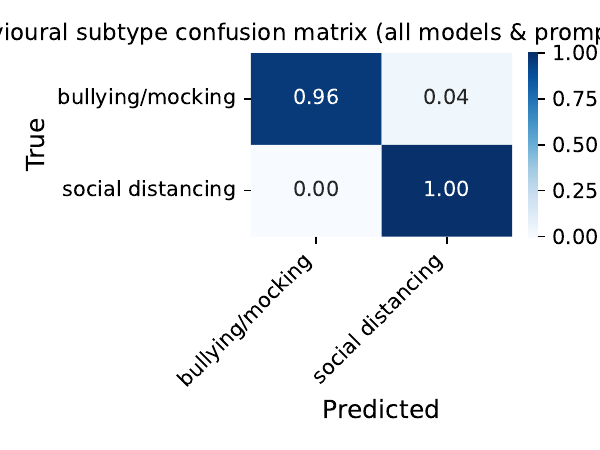}
\caption{Row-normalized model confusion matrix for behavioral public-stigma component prediction.}
\label{fig:model_cm_behavioural}

\end{figure*}

\clearpage

\section{Classifier results for stigma-related constructs}
\label{app:adjacent_constructs}

We evaluated off-the-shelf classifiers for constructs related to mental health stigma: general sentiment, targeted sentiment, hate speech, and toxicity. These models were used as diagnostic baselines rather than as stigma-specific systems. For targeted sentiment, we used the DeBERTa-v3 aspect-based sentiment analysis (ABSA) model to estimate sentiment toward the mental health target term. Although applying an ABSA model outside its conventional aspect-sentiment setting is non-standard, this approach has previously been used to estimate target-directed sentiment in lexical semantic change evaluation and performed well on a synthetic benchmark dataset \citep{baes-etal-2025-lsc}. Table~\ref{tab:adjacent_construct_baselines} lists the models and their intended role in the analysis\footnote{Full model identifiers are: \texttt{yangheng/deberta-v3-base-absa-v1.1}, \texttt{cardiffnlp/twitter-roberta-base-sentiment-latest}, \texttt{cardiffnlp/twitter-roberta-base-hate-latest}, and \texttt{unitary/unbiased-toxic-roberta}.}

\begin{table*}[!htbp]
\centering
\footnotesize
\setlength{\tabcolsep}{4pt}
\renewcommand{\arraystretch}{1.08}
\begin{tabularx}{\textwidth}{@{}p{0.18\textwidth}p{0.30\textwidth}X@{}}
\toprule
\textbf{Construct} & \textbf{Model} & \textbf{Purpose} \\
\midrule
Targeted sentiment & DeBERTa-v3 ABSA & Tests whether stigma is captured by negative sentiment toward the mental health target. \\
General sentiment & Twitter-RoBERTa sentiment & Tests whether stigma is captured by global negative sentiment. \\
Hate speech & Twitter-RoBERTa hate speech & Tests whether stigma overlaps with explicit hate or derogation. \\
Toxicity & Unbiased Toxic RoBERTa & Tests whether stigma overlaps with insult, abuse, or toxic language. \\
\bottomrule
\end{tabularx}
\caption{Neighboring-construct baselines used to test whether mental health stigma is captured by sentiment, hate speech, or toxicity models.}
\label{tab:adjacent_construct_baselines}
\end{table*}

\begin{figure*}[!htbp]
\centering

\includegraphics[width=.72\textwidth]{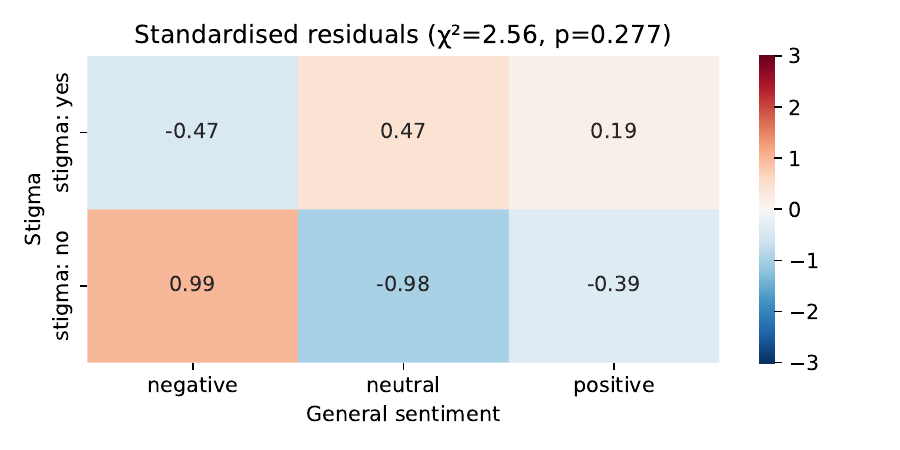}

\vspace{0.7em}

\includegraphics[width=.72\textwidth]{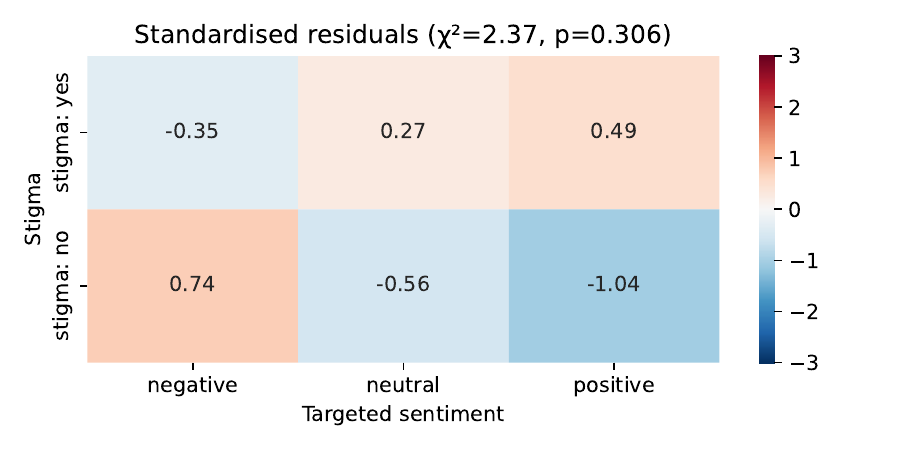}

\vspace{0.7em}

\includegraphics[width=.72\textwidth]{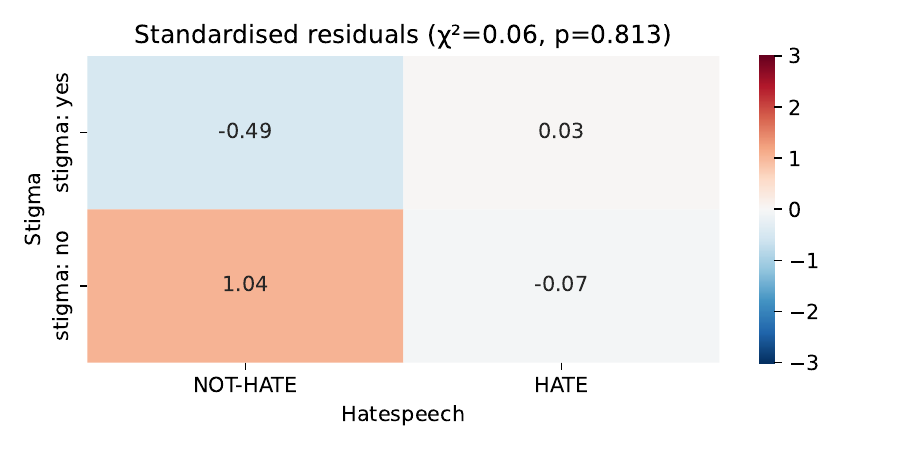}

\caption{Standardized residuals from chi-square tests of independence between gold stigma labels and adjacent classifier predictions. Residuals are shown for predicted general sentiment, targeted sentiment, and hate-speech categories.}
\label{fig:adjacent_residuals}
\end{figure*}

For categorical classifier outputs, we constructed contingency tables crossing the gold binary stigma label with the predicted classifier category and tested their association with chi-square tests of independence. The null hypothesis was that the gold stigma label and predicted classifier category were statistically independent in the population. That is, stigma-present and stigma-absent texts would have the same expected pattern of predicted classifier categories. Table~\ref{tab:adjacent_construct_tests} reports the omnibus tests, and Figure~\ref{fig:adjacent_residuals} shows the corresponding standardized residuals. For toxicity, which produced a continuous score, we inspected score distributions by gold stigma label. Toxicity scores showed substantial overlap between stigma-present and stigma-absent texts and only weak discriminatory ability (AUC = .60), suggesting that toxicity does not provide a reliable proxy for stigma detection.


Across these analyses, classifier outputs for nearby constructs did not map cleanly onto the gold stigma labels. General sentiment, targeted sentiment, and hate-speech predictions were not significantly associated with stigma presence, and toxicity scores showed substantial distributional overlap between stigma-present and stigma-absent texts. These results support the interpretation that mental health stigma is a related but distinct construct, rather than a subset of negative sentiment, hate speech, or toxicity.

\begin{table}[!htbp]
\centering
\small
\begin{tabular}{@{}lccc@{}}
\toprule
\textbf{Classifier prediction} & \textbf{df} & $\boldsymbol{\chi^2}$ & $\boldsymbol{p}$ \\
\midrule
General sentiment & 2 & 2.56 & .277 \\
Targeted sentiment & 2 & 2.37 & .306 \\
Hate speech & 1 & 0.06 & .813 \\
\bottomrule
\end{tabular}
\caption{Chi-square tests of independence between gold stigma labels and classifier predictions for neighboring constructs. Toxicity is not included because it was analyzed as a continuous score rather than a categorical prediction.}
\label{tab:adjacent_construct_tests}
\end{table}

\begin{figure*}[!t]
\centering
\includegraphics[width=.78\textwidth]{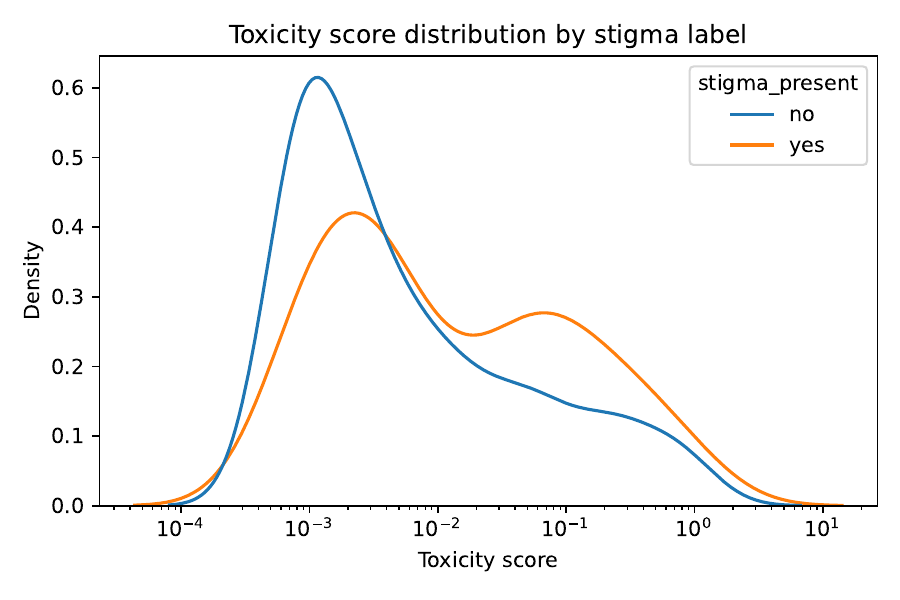}
\caption{Distribution of toxicity scores by gold stigma label. The distributions overlap substantially, suggesting that toxicity scores do not cleanly separate stigma-present from stigma-absent texts.}
\label{fig:toxicity_density}
\end{figure*}

\section{Semantic similarity with stigma prototypes}
\label{app:semantic_similarity}

To estimate if stigma can be captured though superficial semantics in terms of embedding similarity, we compare samples in our dataset with the prototypical examples of stigma we collected \Cref{sec:prototype_exemplars}. Both prototype examples and test samples are encoded using Gemini's gemini-embedding-001. 

For stigma presence detection, all stigma-present prototypes are merged into a single present prototype and all stigma-absent prototypes into a single absent prototype; a text is then classified based on whichever prototype its embedding is closest to. For multilabel fields (mode, domain, subfactor), each text's embedding is compared against the prototypes of all relevant classes within that field and assigned the label of the closest one, producing a single prediction per field. Since this approach is inherently single-label, evaluation is restricted to samples where the gold annotation contains exactly one label for the target field. Prototype embeddings results are then compared against predictions from Gemini 4 Flash. They consistently show significantly worse performance in terms of recall, especially for stigma domain. 
\onecolumn
\begin{figure}[H]
\centering
    \includegraphics[width=0.9\linewidth]{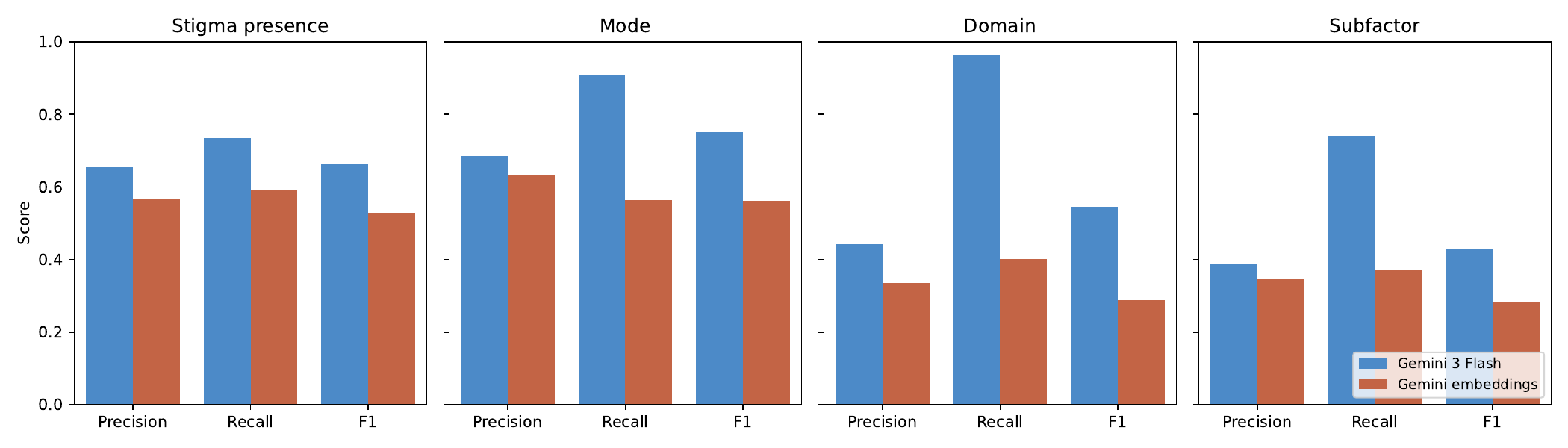}
    \caption{Prediction of stigma presence, mode, domain and subfactor}
    \label{fig:embeddings_comparison}
\end{figure}

\end{document}